\documentclass[preprint,12pt,authoryear]{elsarticle}

\usepackage{amssymb}
\usepackage{amsmath}

\usepackage{bm}
\usepackage{booktabs}

\usepackage{subcaption}   
\usepackage{pdflscape}
\usepackage{adjustbox}
\usepackage{longtable}

\usepackage{xcolor}       
\usepackage{url}     

\DeclareUnicodeCharacter{2061}{} 

\newcommand{\sigmoid}[1]{\varphi\!\left(#1\right)}
\newcommand{\Heav}[1]{\operatorname{H}\!\left(#1\right)}

\newif\ifblindreview
\blindreviewfalse 

\newcommand{\codeavailability}{%
\ifblindreview
A reference implementation of the diffusion normative models (MLP and SAINT denoisers), training scripts, and the full evaluation pipeline used to generate the results and figures in this paper is available at:
\url{https://anonymous.4open.science/r/denoising_diffusion_networks-191E}.
\else
A reference implementation of the diffusion normative models (MLP and SAINT denoisers), training scripts, and the full evaluation pipeline used to generate the results and figures in this paper is available at:
\url{https://github.com/lwhitbread/denoising_diffusion_networks} \citep{whitbread_ddpm_normative_github}.
\fi
}

\journal{Medical Image Analysis}

\begin{document}

\begin{frontmatter}

\title{Denoising diffusion networks for normative modeling in neuroimaging}

\ifblindreview
\else
  \author[aiml,sahmri,scms]{Luke Whitbread\corref{cor1}}
  \ead{luke.whitbread@adelaide.edu.au}
  \author[aiml,sph]{Lyle J. Palmer}
  \ead{lyle.palmer@adelaide.edu.au}
  \author[aiml,sahmri,scms]{Mark Jenkinson}
  \ead{mark.jenkinson@adelaide.edu.au}

  \cortext[cor1]{Corresponding author.}

  \affiliation[aiml]{
    organization={Australian Institute for Machine Learning (AIML), Adelaide University},
    addressline={AIML Building, Lot Fourteen, Cnr North Terrace \& Frome Road},
    city={Adelaide SA 5000},
    country={Australia}
  }
  \affiliation[sahmri]{organization={South Australian Health and Medical Research Institute (SAHMRI)}}
  \affiliation[scms]{organization={School of Computer Science and Information Technology, Adelaide University}}
  \affiliation[sph]{organization={School of Public Health, Adelaide University}}
\fi

\begin{abstract}
Normative modeling estimates reference distributions of biological measures conditional on covariates, enabling centiles and clinically interpretable deviation scores to be derived. Most neuroimaging pipelines fit one model per imaging-derived phenotype (IDP), which scales well but discards multivariate dependence that may encode coordinated patterns. We propose denoising diffusion probabilistic models (DDPMs) as a unified conditional density estimator for tabular IDPs, from which univariate centiles and deviation scores are derived by sampling. We utilise two denoiser backbones: (i) a feature-wise linear modulation (FiLM) conditioned multilayer perceptron (MLP) and (ii) a tabular transformer with feature self-attention and intersample attention (SAINT), conditioning covariates through learned embeddings. We evaluate on a synthetic benchmark with heteroscedastic and multimodal age effects and on UK Biobank FreeSurfer phenotypes, scaling from dimension of 2 to 200. Our evaluation suite includes centile calibration (absolute centile error, empirical coverage, and the probability integral transform), distributional fidelity (Kolmogorov–Smirnov tests), multivariate dependence diagnostics, and nearest-neighbour memorisation analysis. For low dimensions, diffusion models deliver well-calibrated per-IDP outputs comparable to traditional baselines while jointly modeling realistic dependence structure. At higher dimensions, the transformer backbone remains substantially better calibrated than the MLP and better preserves higher-order dependence, enabling scalable joint normative models that remain compatible with standard per-IDP pipelines. These results support diffusion-based normative modeling as a practical route to calibrated multivariate deviation profiles in neuroimaging.
\end{abstract}

\begin{keyword}
Normative modeling \sep Denoising diffusion \sep Conditional density \sep Neuroimaging \sep Transformers \sep Calibration
\end{keyword}



\end{frontmatter}

\section{Introduction}

Normative modeling aims to characterise the distribution of biological measurements in a reference population as a function of independent variables (covariates) such as sex, age, or ethnicity, thereby enabling individual-level deviation scoring that is clinically interpretable \citep{marquand_2016,Wolfers2020,Bethlehem2022}. In neuroimaging, such models map imaging-derived phenotypes (IDPs) against variables such as age and sex and then quantify whether a subject’s measurements depart from the expected normative range for an individual of the same age and sex. Large-scale efforts have underscored the value of this paradigm, for example lifespan ``brain charts'' that provide centiles for diverse structural metrics \citep{Bethlehem2022}. In practice, most implementations adopt a per-phenotype strategy and fit a separate conditional model for each IDP. This univariate formulation yields simple, clinically interpretable centiles and deviation scores for individual measures.

At the same time, many questions in neurodegeneration concern patterns of deviation across multiple IDPs rather than any single measure in isolation \citep{rutherford}. For example, hippocampal atrophy, ventricular enlargement and thinning of Alzheimer’s disease (AD) signature cortices tend to co-occur along disease trajectories \citep{Frisoni2010,Nestor2008,Dickerson2009}; clinicians routinely interpret such combinations as a totality rather than as independent abnormalities. A joint conditional model over vectors of IDPs could in principle support multivariate centiles for anatomically-related sets of regions, composite deviation profiles, and analysis of atypical co-atrophy patterns. From a statistical perspective, a coherent high-dimensional conditional density also ensures that univariate centiles and deviation scores are compatible with the dependence structure observed in the data, instead of being combined post hoc from separate univariate fits. Such a joint model would potentially be both more accurate and more clinically useful.

Building such joint models is technically challenging, largely due to the need to model joint distributions. Classical normative pipelines based on parametric centile regression, Gaussian processes (GPs) or hierarchical Bayesian regression (HBR) have mostly been deployed per phenotype \citep{Bethlehem2022,Boer} and face computational or modelling barriers when extended to dozens or hundreds of IDPs. Recent deep generative approaches --- including VAEs, normalising flows and diffusion/score-based models --- offer a route to learning complex joint densities \citep{Pinaya2021,papamakarios2021,ho,song2021score}, but applications as normative estimators for tabular neuroimaging phenotypes remain comparatively sparse. We briefly review these existing approaches below in Section~\ref{sec:existing-methods}.

\paragraph{This work} We motivate and validate a diffusion architecture for conditional density estimation of IDPs, aimed at constructing normative models from which centiles and calibrated deviation scores can be derived; either as drop-in replacements for established univariate approaches, or as normative estimators for high-dimensional IDP sets. As part of this validation, we experiment with two backbones: (i) an MLP denoiser with FiLM conditioning of covariates \citep{perez2018film}; and (ii) a tabular transformer backbone. For (ii), we instantiate a SAINT-style transformer that applies self-attention across features and inter-sample (row) attention across mini-batches, conditioning covariates via learned embeddings rather than FiLM \citep{saint_trl}. To our knowledge, this is the first application of diffusion modeling to multivariate normative neuroimaging IDPs using a tabular transformer backbone.

\paragraph{Primary contributions}
\begin{itemize}
    \item We adapted denoising diffusion probabilistic models to conditional normative modeling of neuroimaging IDPs, enabling centile and deviation-score estimation from sampled conditional densities $\hat{p}(\mathbf{y} {\mid} \mathbf{c})$.
    \item We compared FiLM-conditioned MLP and SAINT-style transformer denoisers.
    \item We constructed a comprehensive evaluation suite tailored to normative modeling, including centile calibration (absolute centile error, coverage, PIT), distributional tests (Kolmogorov--Smirnov), multivariate dependence diagnostics, and nearest-neighbour memorisation analysis.
\end{itemize}

\subsection{Existing methods}
\label{sec:existing-methods}

\paragraph{Parametric centile charts} LMS and GAMLSS remain mainstays for centile estimation in biomedicine, modeling distributional parameters (e.g., location, scale, skewness) as smooth functions of covariates under flexible parametric families \citep{cole1992lms,rigby2005gamlss}. These frameworks underpin recent MRI ``brain charts'' in which GAMLSS is used to estimate normative trajectories for large numbers of univariate neuroimaging phenotypes across the lifespan \citep{Bethlehem2022}. Strengths include interpretability and mature software. However, parametric centile models can be vulnerable to misspecification in the presence of multimodality or heteroscedasticity, and they do not directly yield a tractable high-dimensional joint density across many IDPs; instead, each IDP is typically modeled separately.

\paragraph{Gaussian processes and hierarchical regression} GP-based normative models capture nonlinear covariate effects and predictive uncertainty \citep{Rasmussen2006}. In practice, exact GPs scale cubically with sample size and are therefore expensive on very large cohorts, even when used in a purely feature-wise manner. Hierarchical Bayesian regression (HBR) extends normative modeling to multi-site neuroimaging with partial pooling of site effects and federated learning; recent work generalises HBR beyond Gaussian errors \citep{Boer}. These approaches are robust and principled for univariate IDPs and small multivariate sets, but multi-output or high-dimensional joint extensions are computationally and statistically challenging, so they are most commonly deployed per phenotype.

\paragraph{Deep generative density models} A range of deep generative architectures can in principle be used for conditional density estimation of continuous or tabular variables. Autoencoder-based latent-variable models (including adversarial autoencoders) have been applied to multi-region neuroimaging measurements to derive reconstruction-error-based deviation scores for normative modelling \citep{Pinaya2021}. Probabilistic variants such as variational autoencoders (VAEs) optimise a variational lower bound (ELBO) rather than the exact marginal likelihood \citep{Kingma2013VAE} and can suffer from failure modes such as posterior collapse, where the decoder ignores part of the latent code \citep{Lucas2019PosteriorCollapse}. VAEs may also produce overly smooth reconstructions under common Gaussian/MSE observation models, particularly in high-dimensional settings \citep{larsen2016vae_gan,gulrajani2016pixelvae}. Normalising flows --- e.g., RealNVP, masked autoregressive flows, and neural spline flows --- use invertible transformations with tractable Jacobians to provide exact likelihoods and flexible conditional densities \citep{RealNVP2016,MAF2017,Durkan2019,papamakarios2021}, but the requirement for global invertibility constrains architectural design and can make expressive flows computationally demanding in high-dimensional settings \citep{papamakarios2021}. Diffusion models (DDPMs and related score-based formulations) avoid explicit invertibility, allow the use of generic denoisers as density estimators, and have demonstrated stable training, strong mode coverage, and excellent sample quality across a range of complex datasets \citep{ho,nichol2021,dhariwal2021,song2021score}. These properties make diffusion models particularly attractive for normative modelling of multivariate neuroimaging IDPs.

\paragraph{Architectures for tabular/high-dimensional IDPs} Attention-based tabular models such as TabNet and FT-Transformer have demonstrated advantages over plain MLPs on mixed-type, high-dimensional data \citep{Arik2021,FTTransformer}. SAINT further adds intersample attention and contrastive pretraining, improving data efficiency and improving performance, including in semi-supervised settings where labels are scarce \citep{saint_trl}. Motivated by these results, in addition to an MLP backbone, we also adopted a SAINT-style transformer with embedding-based conditioning, which we hypothesised would scale to higher-dimensional IDP datasets. We tested this empirically up to dimension $D$ of $200$ IDPs. In the current work we focused on a diffusion-based formulation to evaluate the strengths and weaknesses of MLP and transformer-based backbones for normative density estimation within this framework.


\section{Methods}

\subsection{Data}
\label{subsec:data}

We used two sources of data. First, a synthetic dataset of $N{=}47{,}000$ samples generated from the parametric recipe in section \ref{subsubsec:synthetic_data} (SYNTH). Second, a real-world cohort drawn from the UK Biobank (UKB) imaging study \citep{Miller2016,Bycroft2018}, restricted to participants with T1-weighted (T1w) MRI and complete FreeSurfer‐derived imaging-derived phenotypes (IDPs) \citep{Fischl2012}. For experiments with ${\leq} 20$ IDPs we selected measures with strong priors for neurodegeneration (list below); for higher-dimensional experiments ($D{>}20$) we added randomly sampled additional IDPs to stress-test joint modelling capacity, keeping the neurodegeneration set fixed.

\paragraph{Neurodegeneration IDP set (20 IDPs)} 
We included bilateral hippocampal and amygdalar volumes, bilateral lateral and inferior-lateral ventricular volumes, five bilateral cortical thicknesses (entorhinal, parahippocampal, inferior temporal, middle temporal, posterior cingulate), whole-brain white matter hyperintensity (WMH) volume, and estimated total intracranial volume (eTIV).

\emph{Hippocampus (L/R).} Hippocampal atrophy is a widely used structural MRI marker of Alzheimer’s disease (AD); at the mild cognitive impairment stage it is used as a diagnostic marker and is associated with increased risk of progression to Alzheimer dementia, and it correlates with neuropsychological deficits/cognitive impairment \citep{Frisoni2010}. 

\emph{Amygdala (L/R).} The amygdala shows early atrophy in AD, with losses comparable in magnitude to the hippocampus and correlated with global cognitive severity \citep{Poulin2011}.

\emph{Lateral and inferior-lateral ventricles (L/R).} Ventricular enlargement sensitively indexes global and medial temporal atrophy and relates to longitudinal decline in AD and MCI \citep{Nestor2008}.

\emph{Cortical thickness (bilateral: entorhinal, parahippocampal, inferior/middle temporal, posterior cingulate).} These regions constitute the “cortical signature” of AD; regionally specific thinning in entorhinal/parahippocampal and lateral temporal cortices, as well as the posterior cingulate, tracks symptoms and disease stage \citep{Dickerson2009,Schwarz2016}. 

\emph{WMH volume.} WMHs reflect small-vessel disease burden and contribute to cognitive impairment; greater WMH load is associated with worse outcomes and may modify neurodegenerative trajectories \citep{Debette2010,Prins2015}. 

\emph{eTIV.} Head size is a key normalisation factor for volumetric studies; FreeSurfer’s eTIV provides an atlas-scaling-based proxy for intracranial volume \citep{Buckner2004}, and alternative ICV adjustment strategies can affect effect sizes \citep{Voevodskaya2014}. 

\paragraph{UKB image processing} 
T1w images were processed with the UKB pipeline \citep{AlfaroAlmagro2018} and FreeSurfer \citep{Fischl2012}, yielding \texttt{aseg} subcortical volumes and regional cortical thickness measures based on the Desikan–Killiany (DK) cortical parcellation \citep{Desikan2006}. When building normative models with $D\le 20$, we use the specific IDP fields listed above (Section \ref{subsec:data}); for larger $D$ we augmented with randomly sampled additional FreeSurfer IDPs (subcortical volumes and DK cortical thickness) to probe scalability while preserving clinical interpretability for the core set.

\subsubsection{Synthetic data}
\label{subsubsec:synthetic_data}

We sampled $N{=}47{,}000$ observations using the closed-form mean/variance functions for structures A–D (Eqs.~\ref{eq:muA}–\ref{eq:sigmaD}). Sex was simulated as a balanced binary covariate. The construction induces nonlinearity, heteroscedasticity and (for $x{\ge}65$) bimodality, providing challenging, biologically plausible distributions for calibration tests.

\paragraph{Parametric recipe} Let $X{=}x-65$. For structures A–B, linear/quadratic trends are smoothly blended into alternative polynomial trends via the sigmoid \(\varphi\); structures C--D follow single‑exponential trends across the entire age range (no sigmoid blending). For ages $x{\ge}65$, we introduce a latent subgroup indicator $g{\in}\{-1,+1\}$ with equal probability, and use the corresponding plus/minus sign in the mean functions below. This yields an equal‑weight mixture of Gaussians (skew‑normal for D, shape = 7) beyond age 65.


\begin{align}
\textbf{A:}\quad
\mu_A &= -70X\,\sigmoid{\tfrac{X}{10}} + 20X + 7000
        \;\pm\; \Heav{X}\,\frac{xX}{5} \label{eq:muA}\\
\sigma_A &= 5X\,\sigmoid{\tfrac{X}{10}} + X + 300
           \label{eq:sigmaA}\\[6pt]
\textbf{B:}\quad
\mu_B &= -200(x-15)\,\sigmoid{\tfrac{x-73}{8}} + 45000
        \;\pm\; \Heav{X}\,xX \label{eq:muB}\\
\sigma_B &= 25(x-15)\,\sigmoid{\tfrac{x-73}{8}} + 4500
           \label{eq:sigmaB}\\[6pt]
\textbf{C:}\quad
\mu_C &= 7000\,\mathrm{exp}(-0.04(x-73)) + 25000
        \;\pm\; \Heav{X}\,xX \label{eq:muC}\\
\sigma_C &= 25(x-15)\,\sigmoid{\tfrac{x-73}{8}} + 4500
           \label{eq:sigmaC}\\[6pt]
\textbf{D:}\quad
\mu_D &= 7000\,\mathrm{exp}(0.02(x-50)) + 15000
        \;\pm\; \Heav{X}\,xX \label{eq:muD}\\
\sigma_D &= 7000\,\mathrm{exp}(0.03(x-75)) + 5000
           \label{eq:sigmaD}
\end{align}

In practice, for each synthetic subject with age $x{\ge}65$ we drew $g\in\{-1,+1\}$ with probability $0.5$ and added the corresponding offset term in Eqs.~\ref{eq:muA}–\ref{eq:muD}. For $x{<}65$, the Heaviside function $\Heav{X}$ was zero and the distributions reduced to single-component Gaussians (or skew-normal for D).

\subsubsection{Dataset preprocessing}
\label{subsubsec:dataset_preproc}

All IDPs were z-scaled within the training data (mean 0, unit variance) and the training normalisation was applied to the validation/test splits. For the UKB dataset, we included only participants with complete observations for the chosen IDPs and covariates (age, sex) to avoid imputation confounds in the normative baselines. For UKB we used an 80/20 train/holdout split stratified by age and sex to preserve covariate balance across splits.

\subsection{Normative model construction}
\label{subsec:norm_model_construction}

We sought a conditional normative model for IDPs $\mathbf{y}\in\mathbb{R}^{D}$ given covariates $\mathbf{c}$ (e.g., age, sex):

\begin{equation}
f:\ \mathbf{c}\ \mapsto\ \hat{p}(\mathbf{y} {\mid} \mathbf{c}),
\end{equation}

\noindent from which we derive centiles and joint samples. For univariate summaries of an IDP $y^{(d)}$, the estimated $q$-th centile is $\hat{\zeta}^{(q)}(\mathbf{c}) = \inf\{ t{:}\, \hat{F}_{y^{(d)}\mid\mathbf{c}}(t){\ge} q\}$, with $\hat{F}$ the empirical cumulative distribution function, computed from conditional samples as described in section \ref{subsec:sampling}. In the multivariate setting we used $\hat{p}(\mathbf{y} {\mid} \mathbf{c})$ to draw joint samples and evaluate dependence structure (section \ref{subsec:evaluation}). 

\paragraph{Sampling grid over covariates}
For presentation of centiles and coverage we evaluate on a Cartesian grid spanning $(\text{age}\times\text{sex})$: ages in one-year bins across the observed range and sex $\in\{\text{F},\text{M}\}$. For each grid cell we generated $M$ conditional samples per IDP (and per backbone) and smoothed the centile curves across age with a small $\sigma$ to reduce sampling noise \citep{bozek_2023}. 

\subsubsection{Diffusion model}

We adopted denoising diffusion probabilistic models (DDPMs) for conditional density estimation \citep{ho,nichol2021}. The forward noising process is a fixed Markov chain:

\begin{equation}
    q(\mathbf{y}_{t} {\mid} \mathbf{y}_{t-1})=\mathcal{N}\!\left(\sqrt{1-\beta_{t}}\;\mathbf{y}_{t-1},\; \beta_{t}\mathbf{I}\right),\quad t=1,\dots,T,
\end{equation}

\noindent with $\beta_{t}{\in}(0,1)$ a variance schedule for timestep $t{\in}T$ (we used a linear schedule in this work for simplicity, although note that a number of other variance schedules (e.g., cosine) are commonly used, and reverse-process can be learned to enable faster sampling \citep{nichol2021}) and $\mathbf{y}_{0}\equiv \mathbf{y}$. This Markov chain admits a closed-form $t$-step transition (``one-shot'' noising) distribution:

\begin{equation}
    q(\mathbf{y}_{t} {\mid} \mathbf{y}_{0})=\mathcal{N}\!\left(\sqrt{\bar{\alpha}_{t}}\;\mathbf{y}_{0},\;(1-\bar{\alpha}_{t})\mathbf{I}\right),\quad \bar{\alpha}_{t}=\textstyle\prod_{s=1}^{t}(1-\beta_{s})
\end{equation}

Generation inverts the chain with a denoiser $\boldsymbol{\epsilon}_{\theta}$ conditioned on $\mathbf{c}$:

\begin{equation}
\label{eq:reverse_density}
p_{\theta}(\mathbf{y}_{t-1}\mid \mathbf{y}_{t},\mathbf{c})=\mathcal{N}\!\Bigl(\mu_{\theta}(\mathbf{y}_{t},t,\mathbf{c}),\;\sigma_{t}^{2}\mathbf{I}\Bigr),
\end{equation}

\begin{equation}
\label{eq:reverse_density_mean}
\quad
\mu_{\theta}(\mathbf{y}_{t},t,\mathbf{c})=\frac{1}{\sqrt{1-\beta_{t}}}\left(\mathbf{y}_{t}-\beta_{t}\,\boldsymbol{\epsilon}_{\theta}(\mathbf{y}_{t},t,\mathbf{c})/\sqrt{1-\bar{\alpha}_{t}}\right)
\end{equation}

\noindent where we used the ``predict noise'' parameterisation and fixed $\sigma_{t}^{2}$ as in \citep{ho}. The training objective was the standard noise-prediction loss:

\begin{equation}
\mathcal{L}(\theta)=\mathbb{E}_{\mathbf{y}_{0},\mathbf{c},t,\boldsymbol{\epsilon}}\;\Bigl\|\boldsymbol{\epsilon}-\boldsymbol{\epsilon}_{\theta}\bigl(\sqrt{\bar{\alpha}_{t}}\,\mathbf{y}_{0}+\sqrt{1-\bar{\alpha}_{t}}\,\boldsymbol{\epsilon},\;t,\;\mathbf{c}\bigr)\Bigr\|_{2}^{2}
\end{equation}

\noindent with $t$ sampled uniformly from $T$ and $\boldsymbol{\epsilon}\sim\mathcal{N}(\mathbf{0},\mathbf{I})$.

We conditioned the denoiser on $\mathbf{c}$ using two strategies matched to backbone choice:
\begin{enumerate}
    \item \textbf{FiLM conditioning for MLP.} We employed feature-wise linear modulation (FiLM) layers \citep{perez2018film}, where a covariate MLP maps $\mathbf{c}$ to per-layer affine parameters $(\boldsymbol{\gamma},\boldsymbol{\beta})$ and applies $\text{FiLM}(\mathbf{h})=\boldsymbol{\gamma}\odot \mathbf{h}+\boldsymbol{\beta}$ to hidden activations.

    \item \textbf{Embedding conditioning for SAINT.} For the SAINT backbone we embeded $\mathbf{c}$ and the timestep $t$ into $\mathbb{R}^{d_{\text{model}}}$ via learned linear projections and added these embeddings to every feature token in the sequence, preserving the transformer’s attention structure while injecting covariate information without introducing additional context tokens.
\end{enumerate}

\noindent All models were implemented in PyTorch \citep{Paszke2019}.

\paragraph{Code availability}
\codeavailability

\subsubsection{Backbone architectures}

\paragraph{MLP denoiser}
A fully-connected network with $L$ hidden layers, PReLU activations, dropout and optional batch normalisation. Time $t$ is normalised to $ t \in (0, 1)$ and concatenated to inputs. FiLM layers modulate hidden states by $\mathbf{c}$.

\paragraph{SAINT denoiser}
We adapted a SAINT-style transformer \citep{saint_trl} to serve as the denoiser for continuous IDPs. Each scalar IDP $y^{(d)}$ is mapped to a $d_{\text{model}}$-dimensional ``feature token'' via a learned per-feature weight and bias, plus an additive column embedding, yielding a sequence of $D$ tokens per subject. Covariates $\mathbf{c}$ and the diffusion timestep $t$ are encoded by linear projections into $\mathbb{R}^{d_{\text{model}}}$ and added to every feature token, which conditions the denoiser without increasing the sequence length. The token sequence was then passed through a stack of transformer blocks with self-attention across features (column-wise attention) interleaved with lightweight row-attention blocks. 

Row attention operates on a single summary token per subject (the mean over feature tokens). During training, this summary token was randomly processed in a true intersample mode (we set the random probability of intersample mode processing to 50\% in this work), where multi-head attention is applied across the mini-batch so that each subject’s summary can attend to others in the batch; on other steps, and always at evaluation/sampling time, the same mechanism is used in a ``degenerate'' mode where summaries do not mix across rows. The resulting row-wise update is broadcast back to all feature tokens. This training-only intersample attention acts as a regulariser: it encourages the model to shape its representations using information from similar individuals during optimisation, while ensuring that at test time the learned denoising process for a given subject depends only on that subject’s own IDPs and covariates (i.e., no batch-dependent behaviour). Residual connections, layer normalisation, and feed-forward sublayers follow the standard transformer encoder design; we omitted SAINT’s contrastive pretraining and categorical modules and used only continuous features in our setting.

\subsubsection{Training and sampling}
We used $T$ diffusion steps (100 by default) and a linear $\{\beta_{t}\}$ schedule \citep{ho}. AdamW with gradient clipping is used for optimisation. At test time we drew $M$ conditional samples $\{y^{(m)}\}_{m=1}^{M} {\sim} \hat{p}(\cdot {\mid} \mathbf{c})$ via ancestral sampling; that is, at test time we generated each sample by running the learned reverse Markov chain from pure noise to data (Equations~\ref{eq:reverse_density}--\ref{eq:reverse_density_mean}).

\subsubsection{Baseline model: GAMLSS}

To benchmark the diffusion models against established normative modeling standards, we fit univariate GAMLSS \citep{rigby2005gamlss} for each of the 20-IDP (UKB) and 4-IDP (SYNTH) datasets using the R \texttt{gamlss} package. For each IDP $y^{(d)}$ we used a Sinh--Arcsinh (SHASH) distribution \citep{jones2009sinh}, so that

\begin{equation}
   y^{(d)}_i \sim \mathrm{SHASH}\bigl(\mu_i,\sigma_i,\nu_i,\tau_i\bigr), 
\end{equation}

\noindent where $\mu$ and $\sigma$ are location and scale and $\nu$ and $\tau$ control skewness and tail weight \citep{SHASH}. In the GAMLSS framework each distributional parameter was modeled via its own link function $g_k$ and additive predictor $\eta_{k,i}$ \citep{Stasinopoulos2017GAMLSSBook}:

\begin{equation}
\begin{aligned}
  \eta_{k,i} 
    &= g_k(\theta_{k,i}) \\
    &= \beta_{0,k} + f_{k,\text{cs}}(\text{age}_i;\,\mathrm{df}=3) 
       + \beta_{1,k}\,\text{sex}_i, \\
  &\text{for } k \in \{\mu,\sigma,\nu,\tau\}.
\end{aligned}
\end{equation}

\noindent where $\theta_{k,i}$ denotes the $i$th value of parameter $k$. We used the default link functions for the SHASH family in \texttt{gamlss}, namely identity for $\mu$ and log links for $\sigma,\nu,\tau$ \citep{SHASH}. Models were fitted with a convergence criterion of 0.001 and a maximum of 100 cycles (increased from the default of 20 cycles to improve convergence). If the optimisation reached the maximum number of cycles without satisfying the convergence criterion, or if numerical errors were returned by the optimisation routine, we treated the corresponding IDP model as non-convergent and excluded it from aggregated performance summaries. This occurred for 9 of the 20 UKB IDPs, whereas all SYNTH IDPs converged. For UKB ($D{=}20$), GAMLSS-based results for ACE (Section~\ref{res1:calibration}) and $D_{\text{KS}}$ p-values (Section~\ref{res:distributional_divergence}) are therefore aggregated only over the 11 successfully fitted IDPs.

\subsection{Evaluation}
\label{subsec:evaluation}

The clinical utility of normative models depends on the conditional distribution $\hat{p}(\mathbf{y} {\mid} \mathbf{c})$ supporting: (i) individually interpretable deviation measures, e.g., \textit{this hippocampal volume lies below the 5\textsuperscript{th} centile for a 72-year-old female}, and (ii) reliable estimates of central dispersion, e.g., \textit{a stated 90\% conditional interval actually covers approximately 90\% of healthy subjects in the 72-year-old female cohort}. Hence, we evaluated normative performance in four ways: (i) statistical calibration, (ii) distributional fidelity, (iii) multivariate dependence structure, and (iv) nearest-neighbour memorisation. For multivariate IDPs $\mathbf{y}\in\mathbb{R}^{D}$, we evaluate each marginal $y^{(d)}$ and report joint diagnostics where appropriate.

\subsubsection{Sampling}
\label{subsec:sampling}

Given covariates $\mathbf{c}$, we drew $M$ conditional samples 
$\{\mathbf{y}^{(m)}\}_{m=1}^{M} {\sim} \hat{p}(\cdot{\mid}\mathbf{c})$.
For a univariate marginal $y$ (suppressing the IDP index $(d)$ for clarity) we write $t{\in}\mathbb{R}$ for a threshold in the same units as $y$ (e.g., a hippocampal volume in mm$^{3}$). The empirical conditional CDF (eCDF) at $t$ is the proportion of samples not exceeding $t$:

\begin{equation}
\hat{F}_{y\mid\mathbf{c}}(t)
= \frac{1}{M}\sum_{m=1}^{M}\mathbb{I}\!\left[y^{(m)}\le t\right]
\end{equation}

Intuitively, sweeping $t$ from $-\infty$ to $+\infty$ counts how much probability mass (under the model’s conditional samples) lies at or below each threshold.

The estimated $q$-th centile $\hat{\zeta}^{(q)}(\mathbf{c})$ was defined as the generalised inverse of the eCDF,

\begin{equation}
\hat{\zeta}^{(q)}(\mathbf{c})
\;=\; \inf\{\, t\in\mathbb{R} : \hat{F}_{y\mid\mathbf{c}}(t)\ge q \,\},
\end{equation}

\noindent i.e., the smallest threshold $t$ at which at least a fraction $q$ of samples are ${\le}t$. Equivalently, if $y_{(1)}\le\cdots\le y_{(M)}$ are the order statistics of $\{y^{(m)}\}$, then a simple sample definition is $\hat{\zeta}^{(q)}(\mathbf{c})=y_{(\lceil qM\rceil)}$.

\subsubsection{Evaluation 1: Centile calibration and predictive coverage}
\label{subsec:centile_accuracy}

Following \citep{bozek_2023}, we assessed pointwise centile accuracy; we additionally assess predictive calibration via empirical coverage and the Probability Integral Transform (PIT), whose uniformity is a standard measure of probabilistic calibration in forecast verification and density-forecast evaluation \citep{diebold1998,Gneiting}. 

For these calibration summaries we did not work at the level of individual subjects, but instead averaged over a discrete grid of covariate values $\{\mathbf{c}_{l}\}_{l=1}^{L}$ (e.g., one-year age bins across the observed age range, optionally stratified by sex). Each grid point corresponds to a small ``covariate bin'', within which we pooled held-out subjects whose covariates fell in that region and compared model-based and empirical centiles.

\paragraph{Absolute Centile Error (ACE)} For target quantile $q{\in}(0,1)$ and a covariate grid $\{\mathbf{c}_{l}\}_{l=1}^{L}$ defined above,

\begin{equation}
\mathrm{ACE}(q)=\frac{1}{L}\sum_{l=1}^{L}\left|\hat{\zeta}^{(q)}(\mathbf{c}_{l}) - \zeta_{\text{emp}}^{(q)}(\mathbf{c}_{l})\right|,
\end{equation}

\noindent where $\zeta_{\text{emp}}^{(q)}(\mathbf{c}_{l})$ is the $q$-th quantile computed from held-out observations falling in the covariate bin centred at $\mathbf{c}_{l}$ (we used 1-year age bins). Because all IDPs were standardised to zero mean and unit variance in the training set, ACE is computed in normalised units and can be compared across IDPs. We report mean ACE across covariates and IDPs for each model trained on UKB and SYNTH datasets with MLP and SAINT backbones.

\paragraph{Empirical Coverage Probability (ECP)}\label{sec:coverage} For a nominal interval $(q_{1},q_{2})$ and corresponding central coverage level $a=q_{2}-q_{1}$,
\begin{equation}
\mathrm{ECP}_{q_{1} : q_{2}}=\frac{1}{N}\sum_{i=1}^{N}\mathbb{I}\!\left[y_{i}\in\bigl(\hat{\zeta}^{(q_{1})}(\mathbf{c}_{l}),\ \hat{\zeta}^{(q_{2})}(\mathbf{c}_{l})\bigr)\right],
\end{equation}

\noindent where $y_{i\in N}$ denotes a hold-out sample. We evaluated the reliability of central prediction intervals (PIs) by plotting empirical coverage less nominal coverage across age bins for each IDP. For a nominal level $a$ (e.g., 0.90) we formed central intervals from $\hat\zeta^{(q)}(\mathbf{c}_{l})$ values within each bin and compute the fraction of held-out observations falling inside. A calibrated model will deviate minimally from the nominal level $a$. We show per-IDP deviation curves and a median deviation curve across IDPs to summarise calibration.

\paragraph{Probability integral transform (PIT)} To assess calibration of each conditional marginal, we computed the PIT. For a held-out observation $y$ with covariates $\mathbf{c}$ and fitted conditional CDF $\hat{F}_{y\mid\mathbf{c}}$, define $u{=}\hat{F}_{y\mid\mathbf{c}}(y)$. If the model is well calibrated and $y$ is a continuous outcome, then then for each $\mathbf{c}$ we have $u{\mid}\mathbf{c}{\sim}\mathcal{U}(0,1)$, so a flat PIT histogram indicates good calibration. Systematic ``$\cup$'' or ``$\cap$''-shapes indicate under- or over-dispersion, while sloped or bat-wing shapes indicate bias or tail issues. We compute PIT values within age bins (to respect conditioning) using the eCDF and then pool across bins for stability. 

\subsubsection{Evaluation 2: Distributional divergence}

Coverage, ACE, and PIT target calibration, but do not guarantee that the shape of each conditional marginal (tails, skew, modality) matches the empirical distribution for covariate-matched participants. Hence, we compared the eCDF $\hat{F}_{y\mid\mathbf{c}_{l}}$ to the hold-out CDF $F_{y\mid\mathbf{c}_{l}}$ using the Kolmogorov–Smirnov statistic within each covariate bin:

\begin{equation}
D_{\mathrm{KS}}(\mathbf{c}_{l})=\sup_{t}\bigl|\hat{F}_{y\mid\mathbf{c}_{l}}(t)-F_{y\mid\mathbf{c}_{l}}(t)\bigr|
\end{equation}

Within each bin we obtained a $D_{\mathrm{KS}}(\mathbf{c}_{l})$ p-value in a non-parametric fashion via a label-permutation two-sample test (Real vs Gen), using 500 permutations. Small p-values indicate that the generated conditional samples and held-out data-points in that bin are distinguishable --- i.e., the model's conditional marginal is statistically different from the empirical conditional marginal.

Given that we have many covariate bins and IDPs, we summarise model fit with the fraction of tests whose $D_{\mathrm{KS}}(\mathbf{c}_{l})$ p-value is less than a significance-level (0.05 in this work). A lower fraction is desirable as it indicates that a greater proportion of tests have a generated conditional marginal that is not statistically different from the held-out empirical conditional marginal.

\subsubsection{Evaluation 3. Dependence structure (multivariate)}
\label{subsec:eval_multivariate}

As noted above, normative modeling in neuroimaging is not always about modeling one IDP in isolation. For example, hippocampal atrophy and ventricular enlargement both vary with age and AD-related pathology, and clinical imaging assessment emphasises patterns of atrophy (e.g., medial temporal atrophy alongside ventricular widening); combined indices of hippocampal loss and ventricular expansion can relate more strongly to cognition than either measure alone \citep{ErtenLyons2013,Wahlund2017,Thompson2004}. Hence, we explicitly tested whether $\hat{p}(\mathbf{y}{\mid} \mathbf{c})$ learns realistic multivariate dependencies. We used two complementary analyses: (a) two-sample distances and (b) pair-of-pair comparisons.

\paragraph{Pairwise joints vs product-of-marginals and two-sample distances}
To verify that the model captured dependence beyond reproduction of each marginal separately, we compared generated joint histograms $p(y^{(i)},y^{(j)}\!\mid\!\mathbf{c})$ against a baseline that represents no dependence -- the product-of-generated-marginals  $p(y^{(i)}\!\mid\!\mathbf{c})\,p(y^{(j)}\!\mid\!\mathbf{c})$. The generated joint histograms and real hold-out joint histograms were then compared by computing two well-established two-sample distances between (i) the baseline product-of-generated-marginals and generated joint samples; and (ii) the generated joint samples and the real joint samples for every IDP pair $(i,j)$. The distances used are:

\begin{itemize}
    \item \textbf{Energy distance ($E^{2}$)}, which compares distributions based on expected interpoint Euclidean distances; lower is better \citep{SzekelyRizzo2013}.
    \item \textbf{Maximum Mean Discrepancy ($\text{MMD}^{2}$)} computed with a characteristic kernel (we used an RBF/Gaussian kernel with bandwidth set by the median heuristic); lower is better \citep{Gretton2012}.
\end{itemize}

Both quantities detect distributional mismatches; connections between distance-based (energy) and RKHS/kernel tests (MMD/HSIC) are well-understood \citep{Sejdinovic2013}. We report these distance measures, across all IDP pairs, as distributions where the distances between the real and generated joints can be compared to the distances between the generated joint and the product of generated marginals. If the model is capturing the dependence well then the former distances should be less than the latter, which was visualised using the respective distance distributions.

\paragraph{Higher-order structure via pair-of-pair comparisons}
Capturing only pairwise joints does not guarantee realistic higher-order organisation among different joint distributions. We therefore investigate the similarities between joint distributions for different IDP pairs. We summarise the relationships between all IDP-pairs --- that is, IDPs $(i,j)$ vs $(k,\ell)$ for all pairs $(i,j)$ and $(k,\ell)$ --- by considering joint density shape similarity. For each IDP pair $(i,j)$ we formed a vector by: (i) z-scoring both variables using the sample mean and standard deviation in the dataset (real hold-out or generated), (ii) defining a $B \times B$ grid (where $B{=}15$) with equal-width bins spanning a total range of $[-3,3]$ along each axis, (iii) estimating a 2D  histogram on this grid, and (iv) vectorising the resulting histogram values. 

Correlating these vectors across all pairs yields a symmetric matrix $C_{\text{shape}}$, where each entry is the Pearson correlation between the joint densities (to measure the "shape") of pair $(i,j)$ and pair $(k,\ell)$. Intuitively, each element asks whether the shape of the joint probability density pattern of one IDP pair resembles that of another pair. The overall matrix then provides a visual summary of the structure for which sets of pairs show higher or lower similarities. 

We computed $C_{\text{shape}}$ for Real and Gen, and display both using a shared hierarchical clustering order to effectively impose an ordering of the pairs such that visual structure can be more easily seen. This ordering is calculated using the Real matrix and then the exact same ordering is imposed on the Gen matrix. Specifically, we cluster the Real matrix by agglomerative hierarchical clustering with average linkage (Unweighted Pair Group Method with Arithmetic Mean; UPGMA) on correlation distance $D{=}1-\rho$ and reuse the resulting leaf order for both the Real and Gen matrices, ensuring identical ordering across plots \citep{sokal1958}. If higher order relationships are being captured by the model then the $C_{\text{shape}}$ matrics for Real and Gen should appear very similar.

We also report an absolute-difference heatmap $|C_{\text{shape}}^{\text{Real}}-C_{\text{shape}}^{\text{Gen}}|$ using the same clustering order as above. In addition, to quantify global concordance we applied a Mantel matrix-correlation test (Pearson~$r$ on upper triangles) between the Real and Gen matrices \citep{mantel1967}. This analysis is akin to representational-similarity style comparisons where correlation between structured summaries gauges model–data alignment.

\subsubsection{Evaluation 4: Nearest-neighbour memorisation}

To evaluate training dataset memorisation, we compared, for each generated sample $\mathbf{y}^{(m)}$, the Euclidean distance to its nearest neighbour (NN) in the training set $d_{\text{train}}$ and in the hold-out set $d_{\text{hold}}$. We summarised the ratio $r{=}d_{\text{train}}/d_{\text{hold}}$, where values $r{<}1$ indicate that $y$ is closer to training data than to similar hold out points, consistent with potential memorisation; $r{\approx}1$ is consistent with generalisation. This is a simple, interpretable instance of NN two-sample reasoning used in generative-model evaluation and memorisation studies \citep{Meehan2020,burg2021}; we also report the probability $\text{E}[\mathbb{I}(r{<}1)]$.

\paragraph{Implementation details (nearest-neighbour search)}
To compute $d_{\text{train}}$ and $d_{\text{hold}}$ efficiently we indexed each reference set with a $k$-d tree and answer exact 1-NN queries against that index. $k$-d trees are a classic spatial data structure for nearest-neighbour search that partition $\mathbb{R}^D$ by recursive axis-aligned splits and support fast queries in low–to–moderate dimensions \citep{bentley1979kdtree,friedman1977nearest}. In practice we used an optimised implementation (\texttt{cKDTree}) to build trees for the training and hold-out sets and then query the distance to the closest point for each generated sample \citep{scipy_ckdtree}.

\paragraph{Controlling for sample–size (density) effects}
Because our 80/20 split typically yields a denser training set, a naive NN ratio $r{=}d_{\text{train}}/d_{\text{hold}}$ can be biased towards $r{<}1$ even when the generative model generalises well, since the expected NN distances shrink as sampling density increases. This dependence of NN distances on point density is well-known in spatial statistics (e.g., the expected NN distance decreases with higher point density) \citep{dixon2002nn} and related effects are reflected in nearest-neighbour-based two-sample tests that depend on within-sample NN structure \citep{henze1988}. To remove this density difference, we equalise the sizes of the reference sets by stratified subsampling: we form bins over the covariate space, and within each stratum draw without replacement so that the balanced training and hold-out sets have the same count (the minimum across the two). All NN distances used to calculate $r$ then reference these balanced sets. Our protocol aligns with nearest-neighbour–based evaluation used in recent generative-model memorisation studies \citep{Meehan2020} while explicitly controlling for sample-size effects.


\section{Results}

We report results and summaries for the 20-IDP (UKB) and 4-IDP (SYNTH) datasets using MLP and SAINT backbone models. Additionally, in the Supplementary materials (unless specified below) we report results for models trained on various dimensions (2, 100, and 200 IDPs) and various training set fractions (10\%, 25\%, and 50\%) for UKB.

We also trained univariate GAMLSS models for all IDPs in the UKB (20 IDPs) and SYNTH (4 IDPs) datasets. On the SYNTH dataset, GAMLSS converged for all features. However, on the UKB dataset, the GAMLSS method (SHASH likelihood with cubic splines) exhibited optimisation instability, failing to converge for 9 of the 20 IDPs. Consequently, GAMLSS results reported for ACE (Section \ref{res1:calibration}) and $D_{\text{KS}}$ p-values (Section \ref{res:distributional_divergence}) for UKB ($D=20$) are aggregated only from the 11 successfully fitted IDPs. The diffusion models (MLP and SAINT) successfully trained on all IDPs without stability issues.

Table~\ref{tab:runtimes} reports training time and the time required to generate a fixed batch of $200{,}000$ conditional samples for each model/dataset configuration. On the base 20-IDP UKB runs, the diffusion backbones required a few hundred seconds to train (378 seconds for the MLP and 596 seconds for SAINT), compared with 3569 seconds for the per-IDP GAMLSS baseline. Sampling was efficient for the diffusion MLP (around 9 seconds for all UKB settings) and remained modest for SAINT on low-dimensional models, but became noticeably more expensive as $D$ increased, reflecting the quadratically scaling cost of attention over features. MLP runtimes were largely insensitive to dimensionality (401--451 seconds across all IDP-dimensions), whereas SAINT training and sampling times grew substantially at $D{=}100$ and $D{=}200$. However, it should be noted that the longest training runtime for the 200-IDP SAINT model was still under 2 hours, with sampling time comfortably remaining under half an hour. Both diffusion backbones, however, benefited almost linearly from reductions in the training fraction (Table~\ref{tab:runtimes}), so that learning-curve experiments and smaller reference cohorts remained computationally practical.

\begin{table*}[htbp]
  \centering
  \caption{Training and sampling runtimes (in seconds) for MLP, SAINT, and GAMLSS models on UKB and SYNTH datasets.}
  \label{tab:runtimes}
  \begin{tabular*}{\textwidth}{@{\extracolsep{\fill}} llccrr}
    \toprule
    Architecture & Dataset & Dim.~$D$ & Train frac. & Training (s) & Sampling (s) \\
    \midrule
    \multicolumn{6}{@{\extracolsep{\fill}}l}{\textbf{Base models (full training set)}} \\
    MLP    & UKB   & 20 & 1.00 & 378  & 9   \\
    MLP    & SYNTH & 4  & 1.00 & 159  & 9   \\
    SAINT  & UKB   & 20 & 1.00 & 596  & 125 \\
    SAINT  & SYNTH & 4  & 1.00 & 135  & 22  \\
    GAMLSS & UKB   & 20 & 1.00 & 3569 & 21  \\
    GAMLSS & SYNTH & 4  & 1.00 & 326  & 4   \\
    \midrule
    \multicolumn{6}{@{\extracolsep{\fill}}l}{\textbf{Dimensional scaling (UKB only, full training set)}} \\
    MLP    & UKB   & 2   & 1.00 & 401  & 9   \\
    MLP    & UKB   & 100 & 1.00 & 401  & 9   \\
    MLP    & UKB   & 200 & 1.00 & 451  & 9   \\
    SAINT  & UKB   & 2   & 1.00 & 326  & 15  \\
    SAINT  & UKB   & 100 & 1.00 & 2856 & 575 \\
    SAINT  & UKB   & 200 & 1.00 & 6866 & 1436 \\
    \midrule
    \multicolumn{6}{@{\extracolsep{\fill}}l}{\textbf{Training set fractions / learning curves (UKB only, D=20)}} \\
    MLP    & UKB   & 20 & 0.50 & 183 & 9   \\
    MLP    & UKB   & 20 & 0.25 & 88  & 9   \\
    MLP    & UKB   & 20 & 0.10 & 31  & 9   \\
    SAINT  & UKB   & 20 & 0.50 & 320 & 123 \\
    SAINT  & UKB   & 20 & 0.25 & 146 & 123 \\
    SAINT  & UKB   & 20 & 0.10 & 49  & 122 \\
    \bottomrule
  \end{tabular*}
\end{table*}

\begin{figure*}[htpb]\centering
\includegraphics[width=0.95\textwidth]{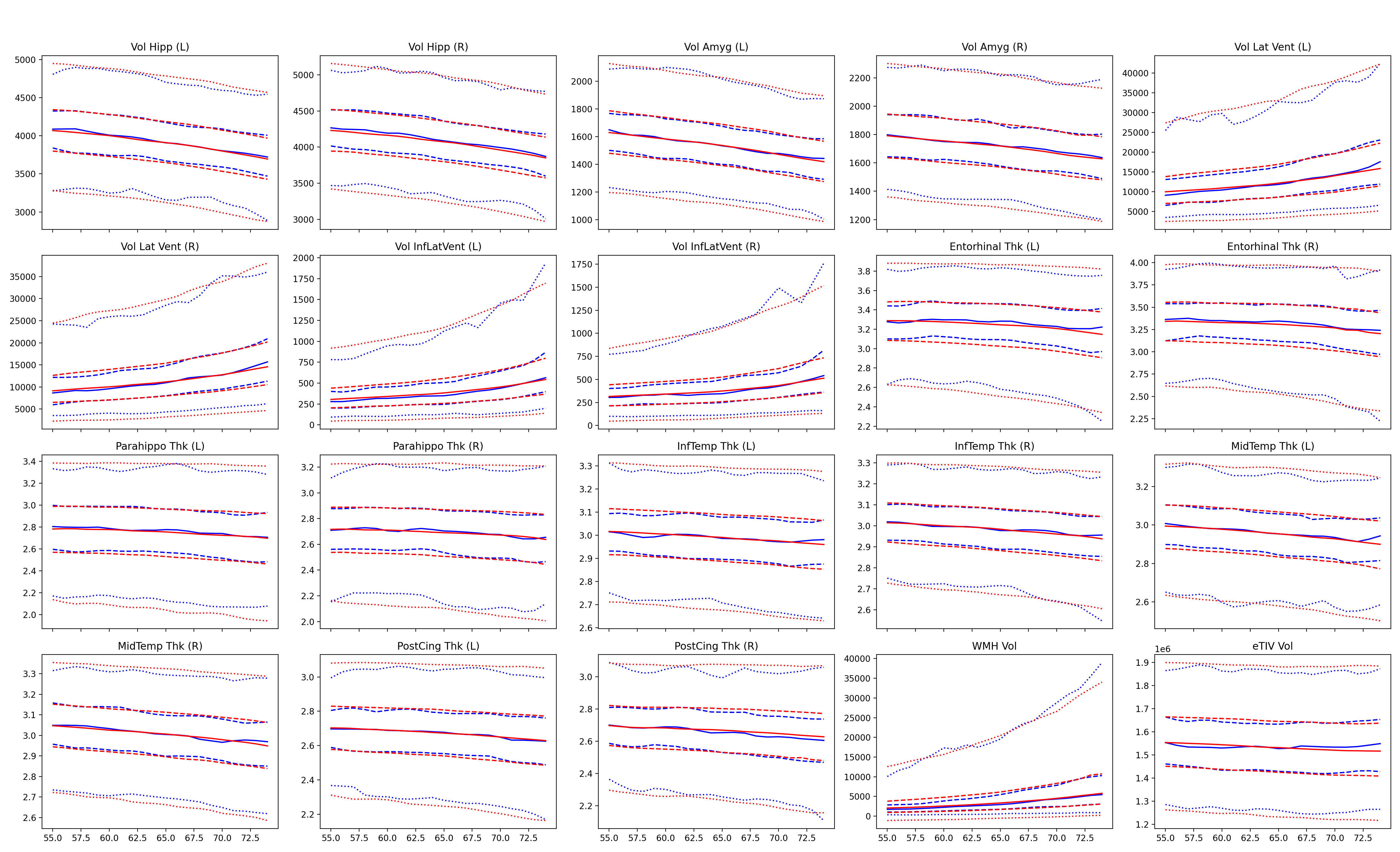}
\caption{Centile curves (base diffusion model): UKB MLP.}
\label{fig:r0_centiles-ukb-mlp}
\end{figure*}

\begin{figure*}[htpb]\centering
\includegraphics[width=0.95\textwidth]{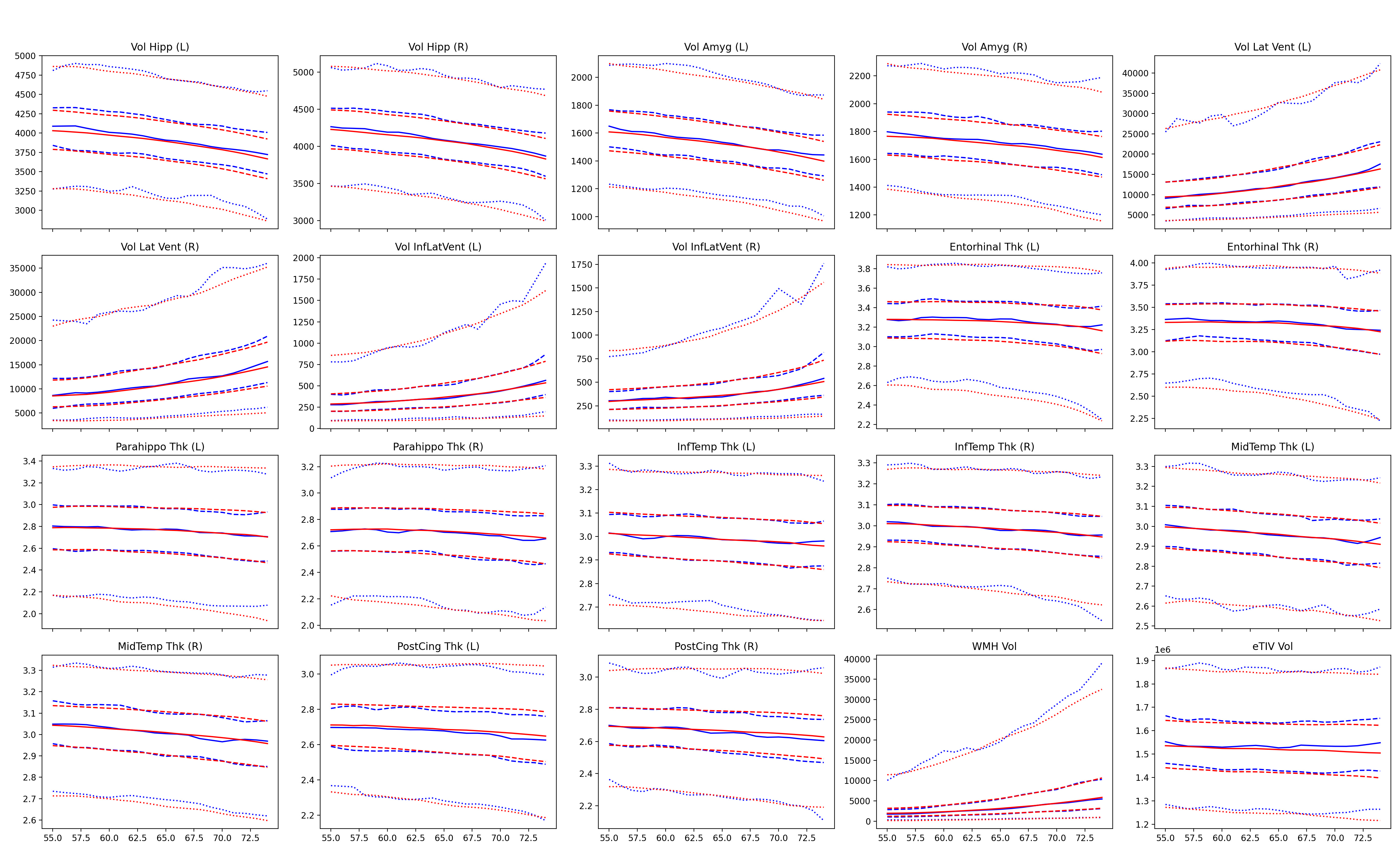}
\caption{Centile curves (base diffusion model): UKB SAINT.}
\label{fig:r0_centiles-ukb-saint}
\end{figure*}

\begin{figure*}[htpb]\centering
\includegraphics[width=0.95\textwidth]{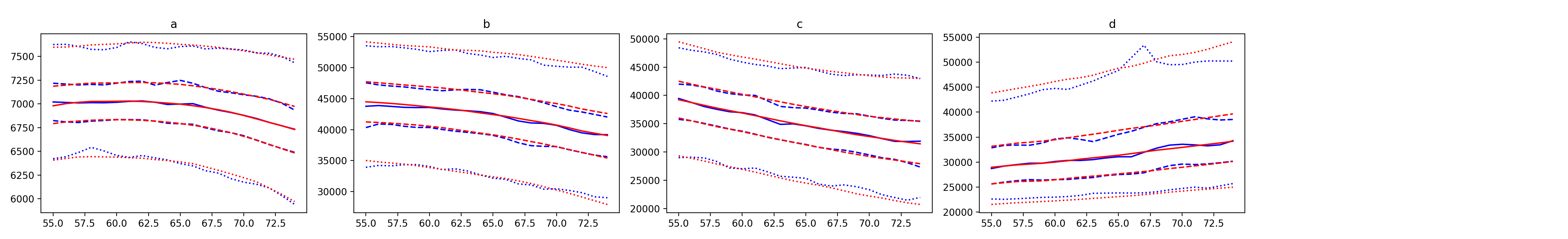}
\caption{Centile curves (base diffusion model): SYNTH MLP.}
\label{fig:r0_centiles-synth-mlp}
\end{figure*}

\begin{figure*}[htpb]\centering
\includegraphics[width=0.95\textwidth]{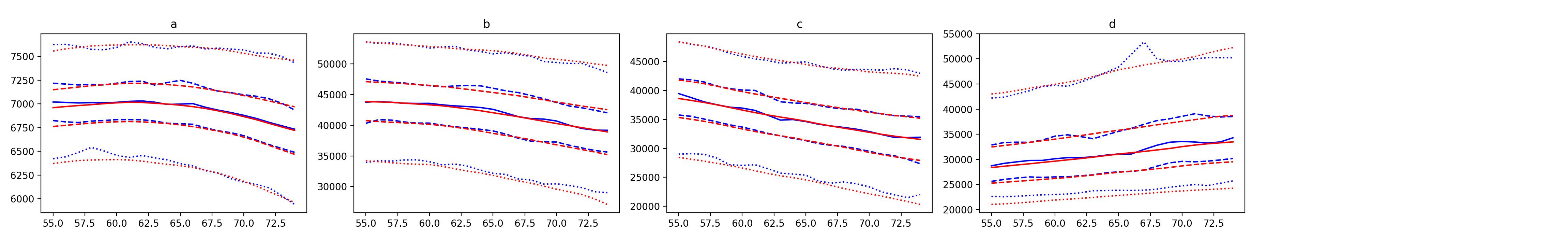}
\caption{Centile curves (base diffusion model): SYNTH SAINT.}
\label{fig:r0_centiles-synth-saint}
\end{figure*}

Figures~\ref{fig:r0_centiles-ukb-mlp}--\ref{fig:r0_centiles-synth-saint} report centile curves vs age for the 20-IDP (UKB) and 4-IDP (SYNTH) with both MLP and SAINT models. Across UKB (D=20), visual inspection of the centile trajectories (Figures~\ref{fig:r0_centiles-ukb-mlp}--\ref{fig:r0_centiles-synth-saint}) shows plausible age trends across neurodegeneration-sensitive IDPs (e.g., hippocampus, ventricles, AD-signature cortices).

\subsection{Results 1: Centile calibration and predictive coverage}
\label{res1:calibration}

\begin{figure*}[htpb]\centering
\includegraphics[width=0.95\textwidth]{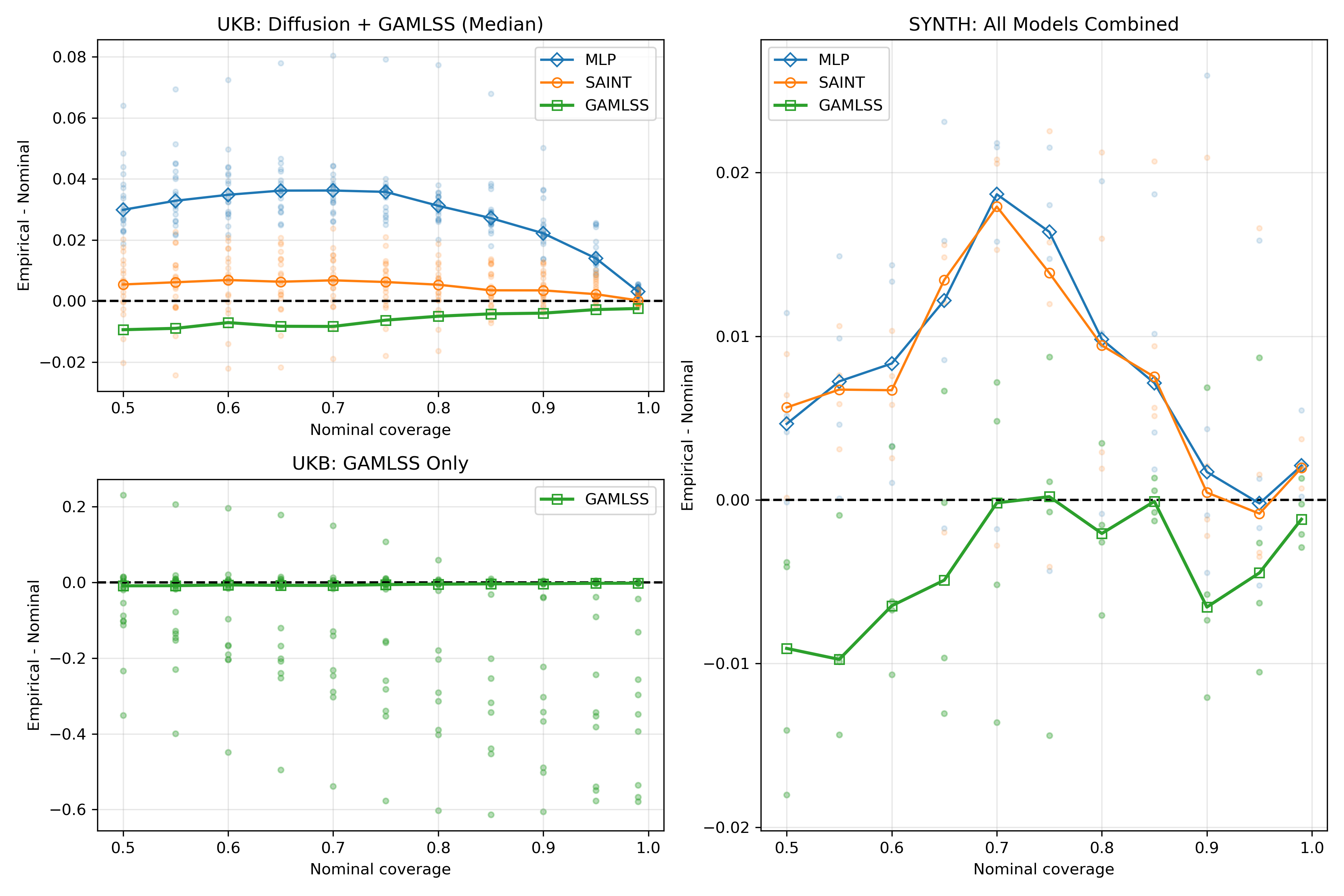}
\caption{Coverage difference overlays for base runs: left column shows UKB (GAMLSS median shown on top row to compare to diffusion-based MLP and SAINT, with the individual data points for GAMLSS shown on the bottom row, while corresponding individual data points for the diffusion-based models are shown on the top row); right column shows medians and individual data points in SYNTH for GAMLSS,  diffusion-based MLP and SAINT.}
\label{fig:r1_covdiff_overlay}
\end{figure*}

\paragraph{Reporting} For central $(100a)\%$ prediction intervals we plot the difference between empirical coverage and the nominal level $a$, by age bin and IDP (as points). Median values, across IDPs, are shown as curves to summarise per-backbone calibration. Figure~\ref{fig:r1_covdiff_overlay} shows these median curves for the 20-IDP (UKB) and 4-IDP (SYNTH) datasets with the MLP and SAINT backbones for diffusion models and the per-IDP GAMLSS model. 

Across UKB (D{=}20), both diffusion backbones produce well-calibrated central PIs, with SAINT’s median coverage curve tracking the identity more tightly than MLP (Figure~\ref{fig:r1_covdiff_overlay}); and while the median GAMLSS central PI was well-calibrated, the results clearly demonstrate failures on a number of IDPs with the absolute difference between empirical coverage and the nominal level $a$ greater than 0.6 in some instances (Figure~\ref{fig:r1_covdiff_overlay}, bottom left). On SYNTH (D=4), all models were close to nominal.

\begin{figure*}[htpb]\centering
\includegraphics[width=0.9\textwidth]{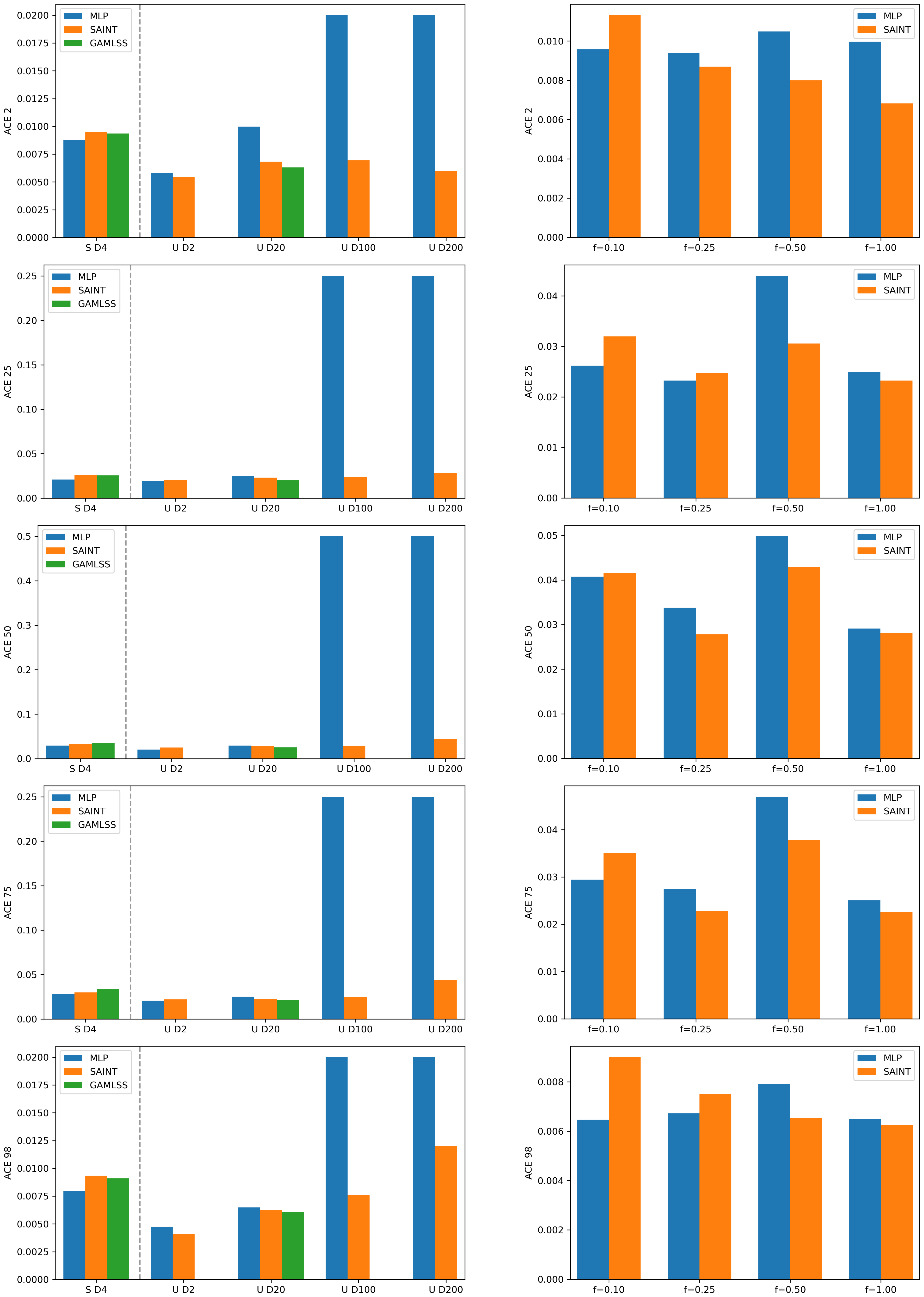}
\caption{ACE bar plots (ACE 2/25/50/75/98): left = dimensional scaling (SYNTH + UKB) with additional GAMLSS bars for S D4 and U D20 (successful GAMLSS IDPs only); right = training fractions (UKB, diffusion models only).}
\label{fig:r1_ace}
\end{figure*}

Figure~\ref{fig:r1_ace} reports mean ACE across IDPs for diffusion models trained on various dimensions of IDPs and training dataset fractions for the baseline 20-IDP UKB dataset, as well as for SYNTH dataset. We also compute the probability integral transform (PIT) $u=\hat F_{y\mid\mathbf{c}}(y)$ per age bin and pool across bins. For calibrated conditionals, PIT is $\mathcal{U}(0,1)$; thus PIT histograms should be flat \citep{diebold1998}.

Quantitatively, the GAMLSS baseline achieved competitive ACE scores on the UKB dataset (D=20) for the subset of IDPs where it converged, and performed similarly to the diffusion backbones on the SYNTH dataset (Figure~\ref{fig:r1_ace}). SAINT attained similar mean ACE to MLP on UKB and SYNTH in lower dimensions and higher training dataset fraction regimes; (Figure~\ref{fig:r1_ace}). In high-dimensional settings (D{=}100/200), MLP degraded sharply --- ACE results saturated at their respective centile level --- whereas SAINT remained relatively well-calibrated. The observed pattern aligns with expectations for attention-based tabular denoisers in high-dimensional settings \citep{saint_trl}.

\begin{figure*}[htpb]\centering
\includegraphics[width=0.95\textwidth]{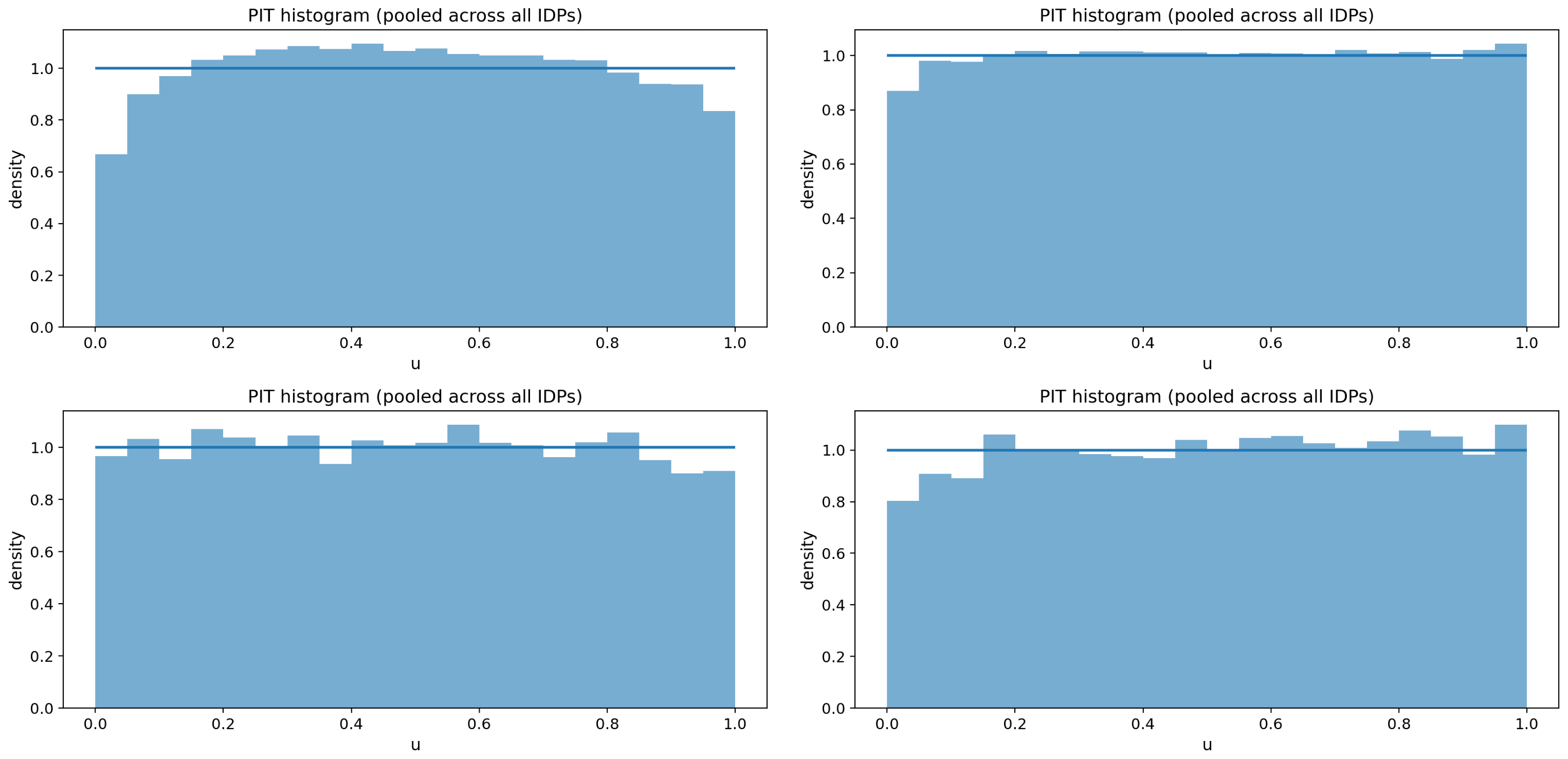}
\caption{Pooled PIT histograms (base diffusion models): UKB and SYNTH for MLP/SAINT.}
\label{fig:r1_pit}
\end{figure*}

Figure~\ref{fig:r1_pit} shows pooled PIT histograms for the MLP and SAINT backbones trained on the 20-IDP UKB dataset and the 4-IDP SYNTH dataset. Additional plots at other dimensionalities and training fractions are provided in the Supplementary material.

Pooled PIT histograms were approximately uniform for UKB (D=20) with both backbones (Figure~\ref{fig:r1_pit}), indicating calibrated conditional marginals with only mild bias observed (subtle $\cap$-shape for the MLP). On SYNTH (D=4), PIT histograms were again well-calibrated. Under high-dimension regimes (Supplementary material), the MLP backbone exhibited poor PIT performance with mass near $u {\approx} 0.5$, while the SAINT backbone exhibited more robust PIT performance, although bias (sloped PIT histograms) was observed for 200-IDP model.

\subsection{Results 2: Distributional divergence}
\label{res:distributional_divergence}

\begin{figure*}[htpb]\centering
\includegraphics[width=0.52\textwidth]{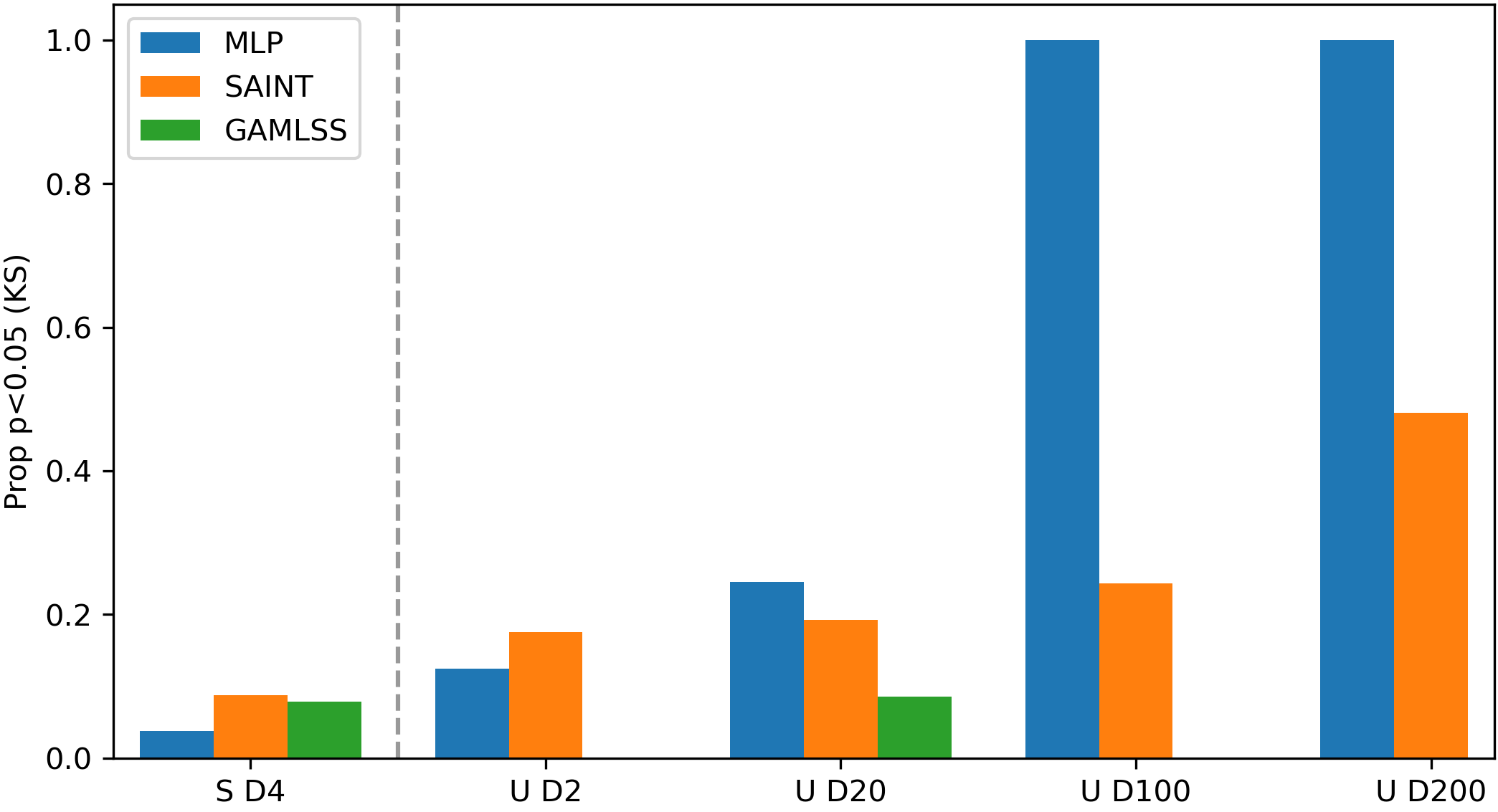}
\hfill\includegraphics[width=0.45\textwidth]{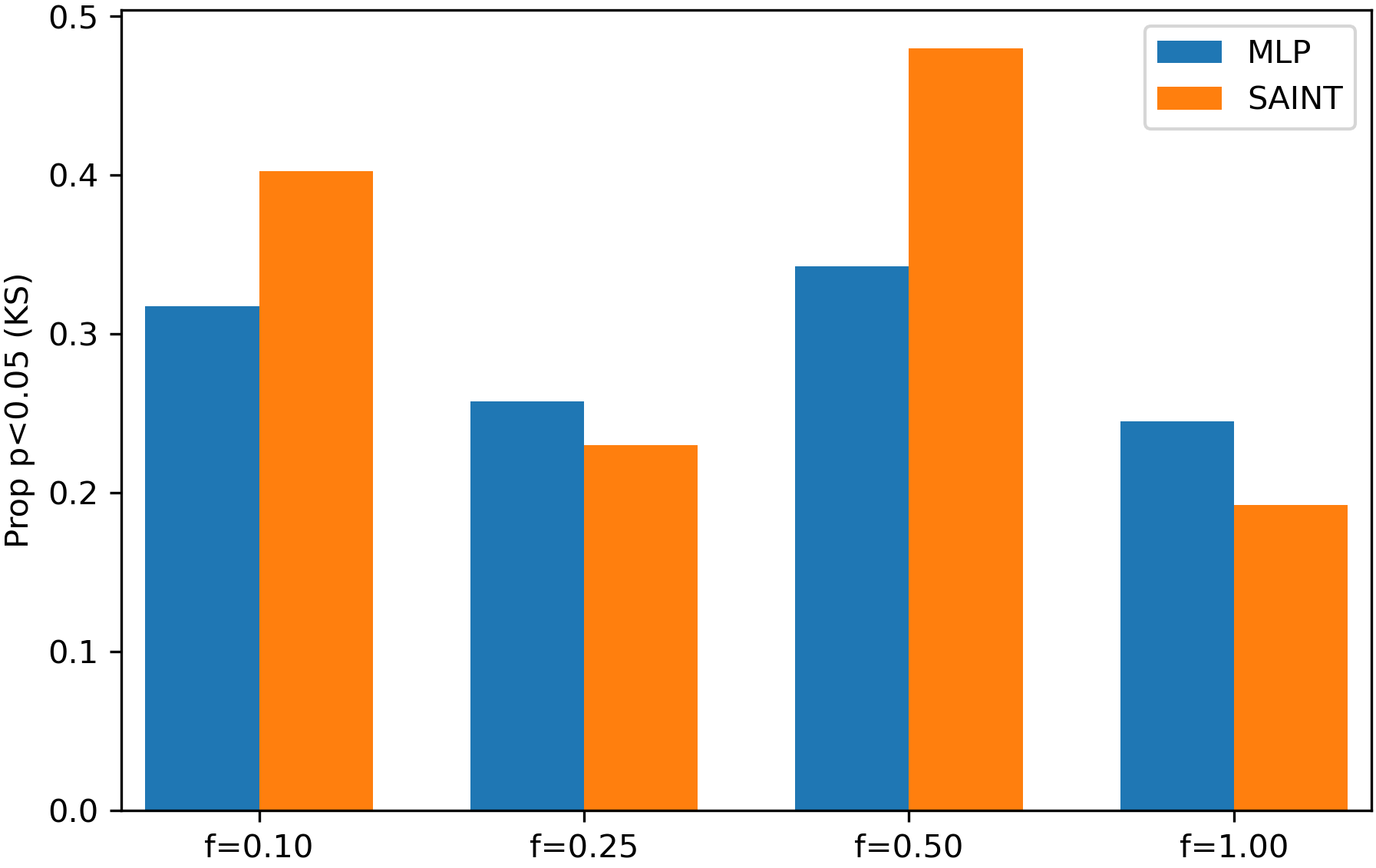}
\caption{Proportion of raw KS p-values below 0.05: Dimensional scaling (SYNTH and UKB) panel (left) shows diffusion models (MLP/SAINT) with additional GAMLSS bars for SYNTH D4 and UKB D20 (successful GAMLSS IDPs only); training fractions (UKB) panel (right) shows diffusion models only. Lower is better.}
\label{fig:r2_ks}
\end{figure*}

\paragraph{Reporting} We report (Figure~\ref{fig:r2_ks}) the proportion of ${D}_{\mathrm{KS}}$ p-values that are less than 0.05 across covariate bins and IDPs for UKB and SYNTH for both diffusion backbones (MLP and SAINT) and the GAMLSS baseline. Note that for UKB (D=20), the GAMLSS bar represents the rejection fraction only for the subset of converged models. Lower bars represent better performance of the model.

On UKB, GAMLSS generally exhibited low rejection rates for the successfully modeled IDPs, comparable to the diffusion models in the $D=20$ regime, meaning most marginals from the generative model were not distinguishable from held-out data under ${D}_{\mathrm{KS}}$. On SYNTH (4-IDP), fractions were similarly low, which is unsurprising given that the SYNTH generative process is relatively simple.

However, in higher-dimensional regimes (i.e., $D{=}100$ and $D{=}200$) on UKB, the MLP backbone’s rejection fraction saturated, indicating that its per-IDP conditional marginals drift away from those observed in held-out subjects; this was consistent with the failure modes seen in centile accuracy, coverage, and PIT. By contrast, SAINT’s rejection fraction remained bounded in these high-dimensional regimes, supporting the claim that SAINT learns realistic conditional marginals even in very high dimensional joint models.

\subsection{Results 3: Dependence structure (multivariate)}
\label{sec:dep-structure}

\begin{figure*}[htpb]\centering
\includegraphics[width=0.95\textwidth]{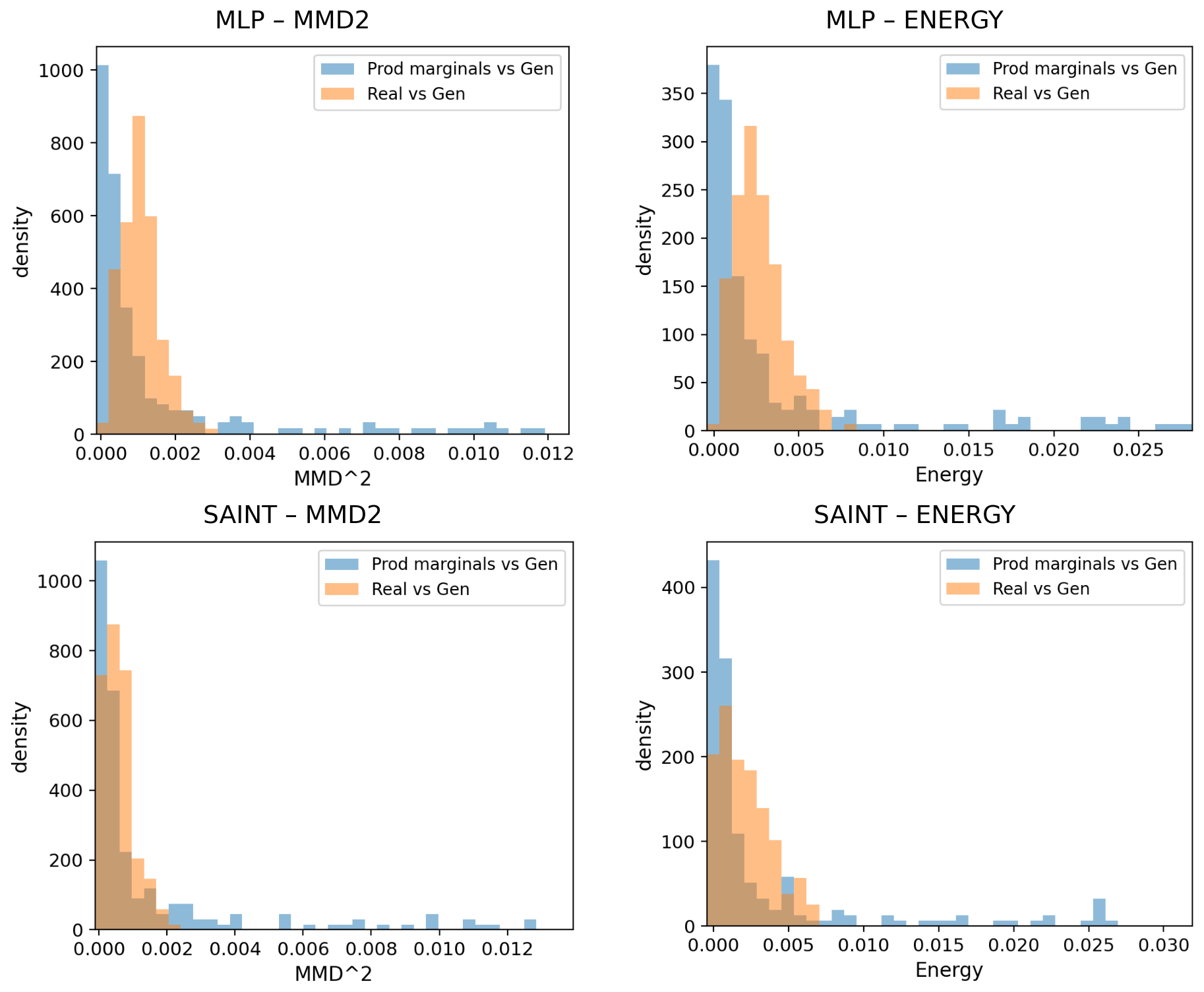}
\caption{Distribution overlays for MMD$^2$ and Energy (diffusion models): UKB base MLP and SAINT.}
\label{fig:r3_stats}
\end{figure*}

\begin{figure*}[htpb]\centering
\includegraphics[width=0.95\textwidth]{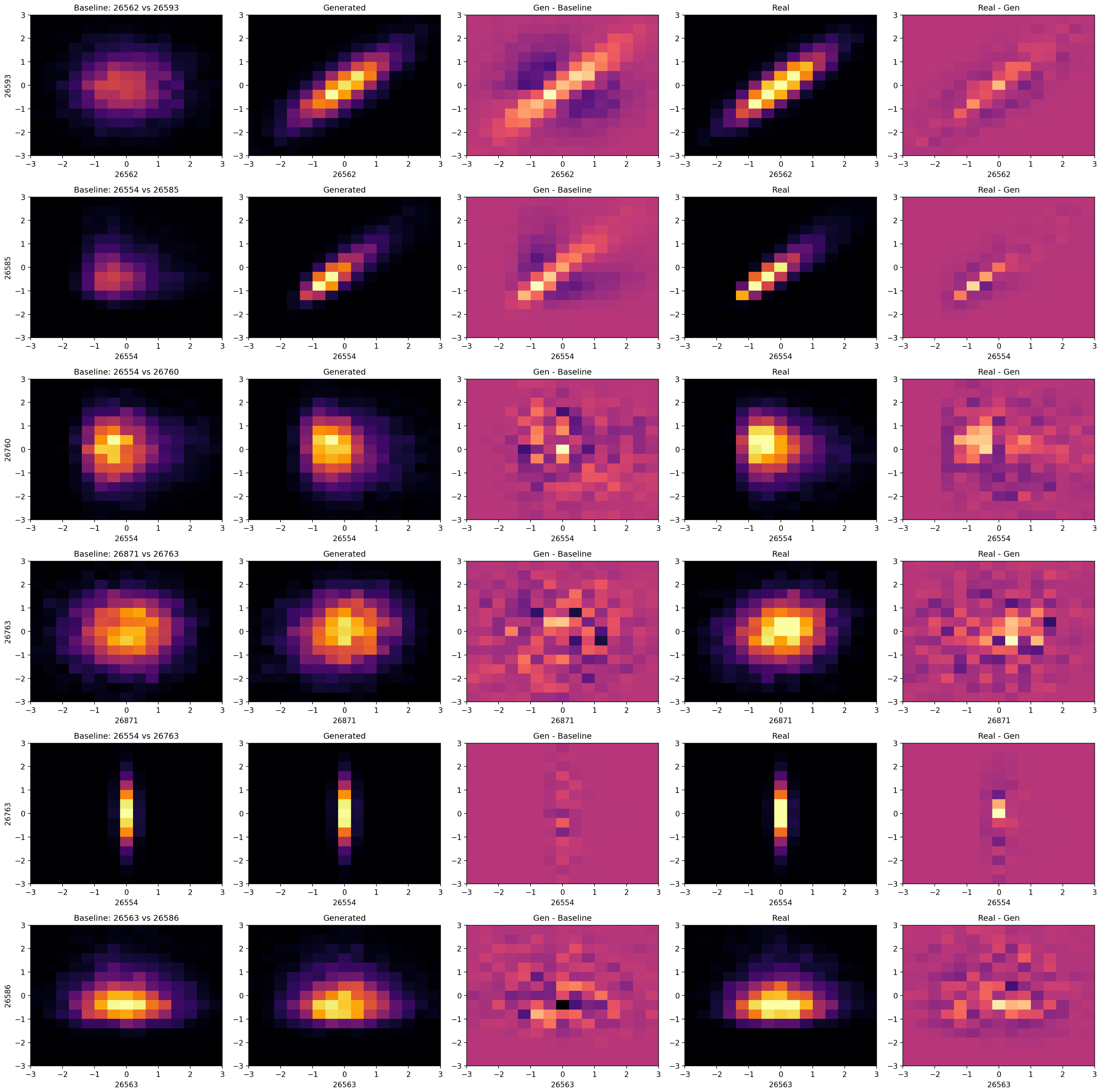}
\caption{Ranked joint pairs (top/middle/bottom k=2) by MMD$^2$ for UKB base MLP diffusion models.}
\label{fig:r3_ranked_mlp}
\end{figure*}

\begin{figure*}[htpb]\centering
\includegraphics[width=0.95\textwidth]{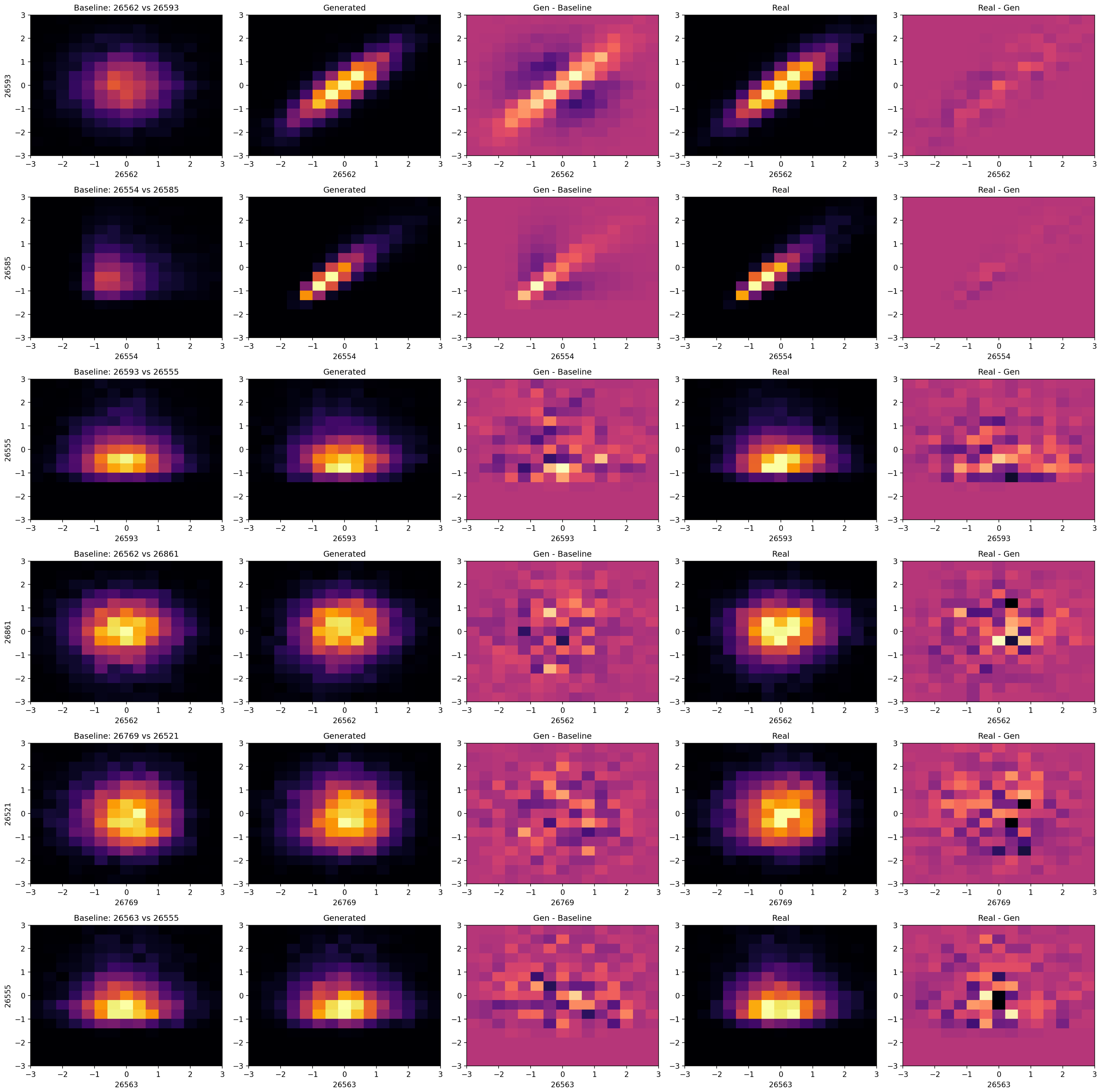}
\caption{Ranked joint pairs (top/middle/bottom k=2) by MMD$^2$ for UKB base SAINT diffusion models.}
\label{fig:r3_ranked_saint}
\end{figure*}

\begin{figure*}[htpb]\centering
\includegraphics[width=0.95\textwidth]{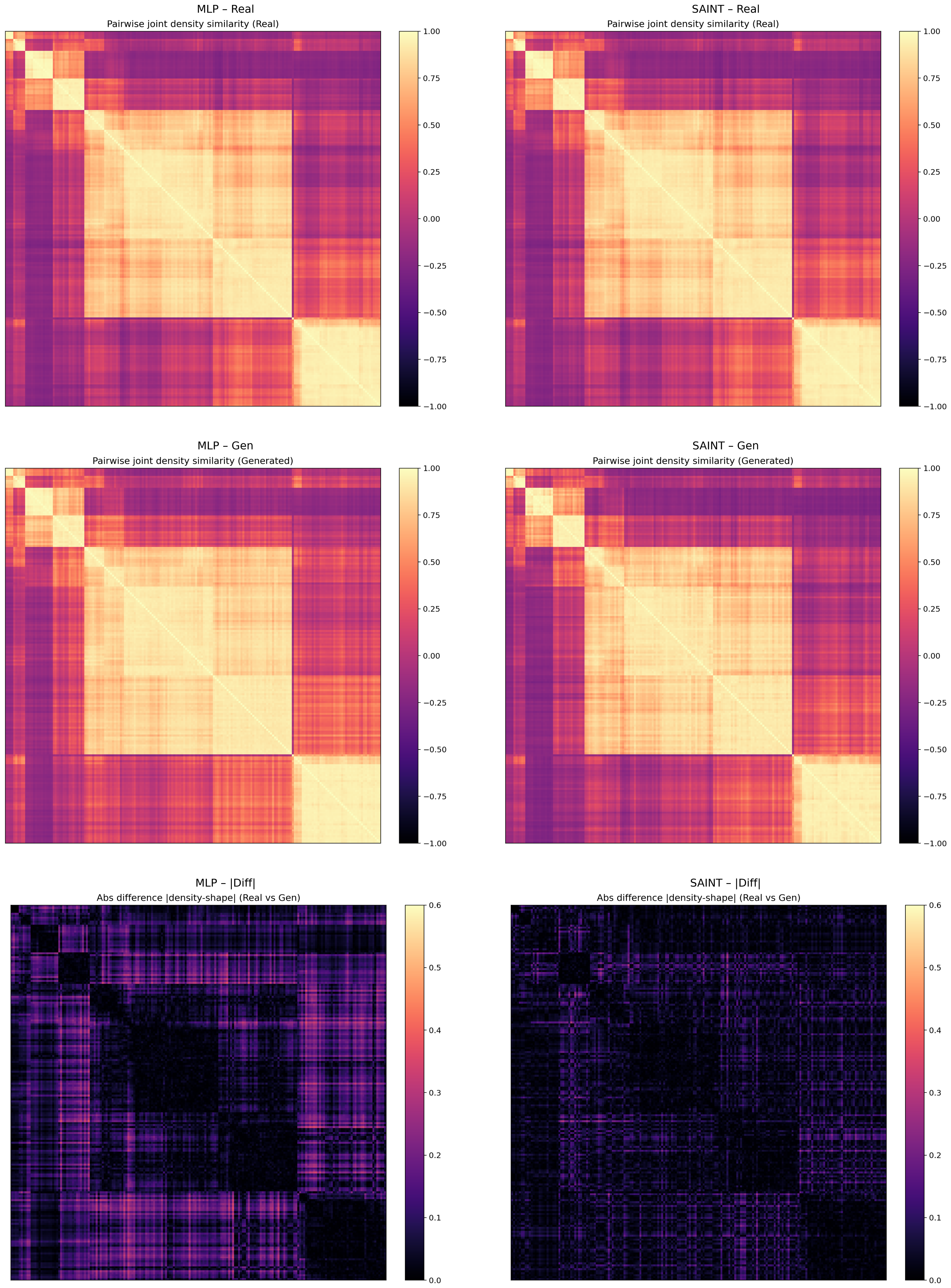}
\caption{Pair-of-pair density-shape correlation (diffusion models): Real, Gen, and |Diff| for UKB base MLP and SAINT.}
\label{fig:r3_pair_of_pair}
\end{figure*}

\paragraph{Reporting} We provide distributions for $E^2$ and MMD$^2$ distances between the product-of-generated-marginals and generated joint histograms, and between the generated joint histograms and the real joint histograms (for 20-IDP UKB models; Figure~\ref{fig:r3_stats}). In addition, we provide panels showing examples of joint distributions for: (i) a baseline of the product-of-generated-marginals $p(y^{(i)}\!\mid\!\mathbf{c})\,p(y^{(j)}\!\mid\!\mathbf{c})$; (ii) generated joint histograms $p(y^{(i)},y^{(j)}\!\mid\!\mathbf{c})$; (iii) a difference map between the baseline and generated joint marginals; (iv) real joint histograms; (v) and a difference map between generated and real joint histograms (we show the best 2, middle 2, and worst 2 pairs panels, ranked by MMD$^2$ computed between the product-of-generated-marginals and generated joint histograms; Figures~\ref{fig:r3_ranked_mlp}--\ref{fig:r3_ranked_saint}). Finally, we show heatmaps of $C_{\text{shape}}$ for Real and Gen (shared clustering order) and absolute-difference maps (Figure~\ref{fig:r3_pair_of_pair}).

When the two-sample distances show that generated joint distributions are far from the product-of-generated-marginals, but close to the corresponding real joint distributions, this suggests that the model is capturing genuine dependence rather than just the marginals. In addition, the strong similarity between the Real and Gen pair-of-pair shape-correlation matrices, together with small absolute-difference maps, indicates that the diffusion model also recovers the higher-order arrangement of dependencies across IDP pairs.

\paragraph{Pairwise structure} For UKB (D=20), generated pairwise joints visibly depart from the product-of-generated-marginals baseline for the best k=2 IDP pairs, more closely aligning with joint density shapes in the held-out data (Figures~\ref{fig:r3_ranked_mlp}-- \ref{fig:r3_ranked_saint}). For each IDP pair we computed a two-sample distance (Energy, MMD) between the generated joint and the real joint, and then took the median of these distances across all pairs. The median distances are generally smaller for SAINT than for the MLP backbone, indicating closer alignment of SAINT's learned pairwise joints with the empirical distributions.

\paragraph{Pair-of-pair structure} The pair-of-pair “shape” correlation matrices $C_{\text{shape}}$ showed strong concordance between Real and Gen, with SAINT slightly closer (smaller absolute difference maps in Figure~\ref{fig:r3_pair_of_pair}). Mantel correlations summarise this: 0.985 for MLP and 0.991 for SAINT, consistent with both MLP and SAINT preserving not just individual pairwise shapes but also their arrangement across IDP-pairs.

\subsection{Results 4: Nearest-neighbour memorisation}
\label{sec:nn-mem}

\begin{figure*}[htpb]\centering
\includegraphics[width=0.95\textwidth]{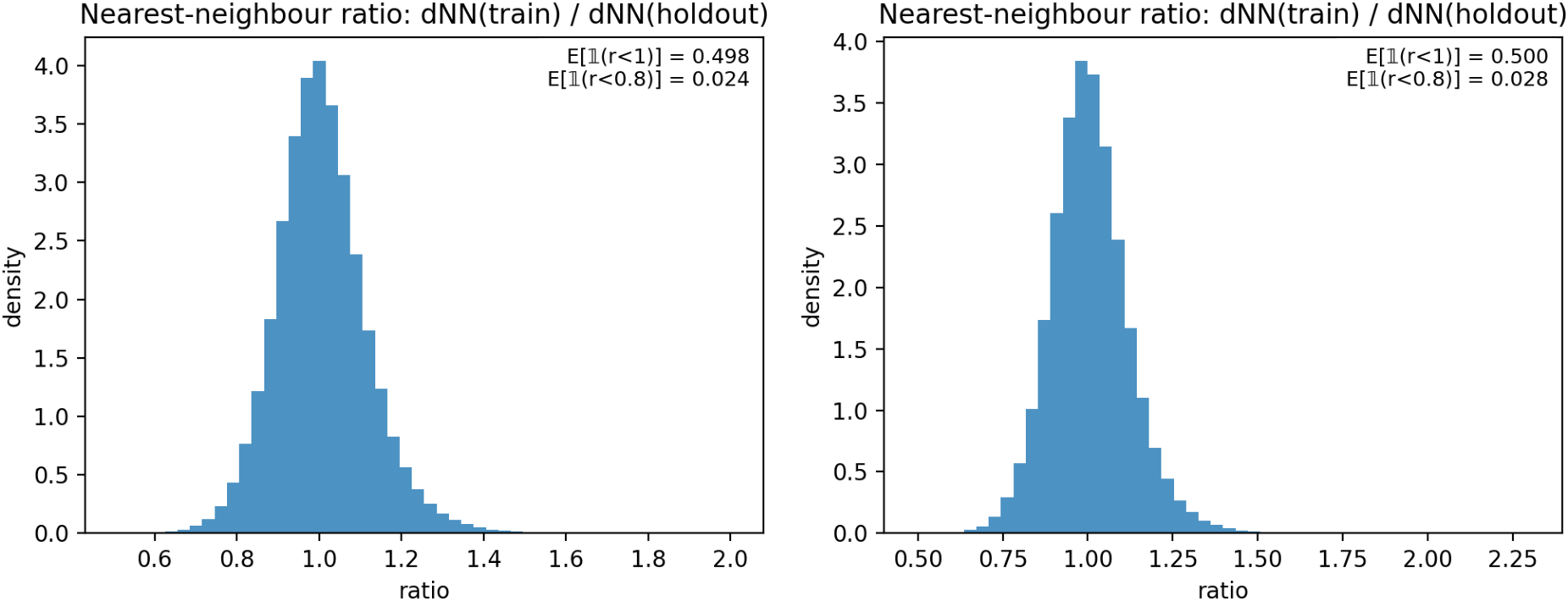}
\caption{Nearest-neighbour ratio $r$ histograms for base UKB diffusion runs: MLP and SAINT.}
\label{fig:r4_nn}
\end{figure*}

\paragraph{Reporting} We show histograms of $r$ values for the MLP and SAINT backbones trained on the 20-IDP set for the UKB data (Figure~\ref{fig:r4_nn}). For each generated sample we computed distances to the nearest neighbour in train and hold-out datasets, $d_{\mathrm{train}}$ and $d_{\mathrm{hold}}$, and summarised the ratio $r=d_{\mathrm{train}}/d_{\mathrm{hold}}$. Values near~1 indicate generalisation; left-shifted mass ($r<1$) suggests potential memorisation.

After balancing the training and hold-out datasets by covariate strata, the NN-ratio distributions were centred at $r=1$ for both backbones ($\text{E}[\mathbb{I}(r{<}1)]$ of 0.498 and 0.500 for MLP and SAINT, respectively). The observed behaviour is what would be expected under the null hypothesis of exchangeable labels when the two reference datasets (train vs hold-out) are matched in size and covariate distribution; that is, a generated point is equally likely to have its nearest neighbour in either set, providing evidence that the models have effectively learned the underlying conditional data distribution. 


\section{Discussion}
\label{sec:discussion}

We have demonstrated that denoising diffusion probabilistic models provide an effective and flexible framework for constructing normative models from high-dimensional neuroimaging data. Our results establish that diffusion models can simultaneously deliver calibrated centile estimates comparable to traditional parametric methods while naturally capturing multivariate dependence structure across imaging-derived phenotypes. This work represents the first application of diffusion models with tabular transformer backbones to normative modeling in neuroimaging, and our evaluation framework provides empirical evidence that these models scale effectively to higher-dimensional joint distributions that challenge conventional approaches.

\subsection{Diffusion models as normative estimators}

Diffusion models offer several advantages as normative estimators that address longstanding challenges in neuroimaging. Traditional distributional regression methods such as LMS and GAMLSS specify a parametric family; multimodality can be difficult to capture with a single parametric family (often motivating mixture modeling extensions), and tail behaviour depends on selecting an appropriate distributional form (and link functions) for the application \citep{cole1992lms,rigby2005gamlss,dinga2021normative}. By contrast, diffusion models learn the conditional density $\hat{p}(\mathbf{y}{\mid}\mathbf{c})$ through iterative denoising in a comparatively unconstrained manner. This flexibility is attractive for neuroimaging data, where IDPs frequently exhibit complex, non-Gaussian conditional distributions shaped by biological heterogeneity, measurement noise, and batch (e.g., scanner) effects \citep{rutherford}.

Diffusion models have been shown empirically to provide strong mode coverage and stable training via a simple denoising objective \citep{song2021score,nichol2021,dhariwal2021}. In a normative modeling context, these properties support the requirement to approximate potentially complex conditional densities $p(\mathbf{y}{\mid}\mathbf{c})$ rather than enforcing a single parametric form. Our synthetic experiments (Section~\ref{subsubsec:synthetic_data}) make this explicit: the age-dependent mixture construction yields bimodal conditional distributions, and the diffusion-based normative models allocate probability mass to both modes in a way that is consistent with the empirical centiles and KS-based distributional tests (Sections~\ref{res1:calibration} and~\ref{res:distributional_divergence}). Moreover, the calibration and KS results demonstrate that the denoising formulation can recover multi-modal, heteroscedastic conditionals that would be difficult to capture with a single parametric family.

The same pattern is seen on UKB data. Both MLP and SAINT backbones yield conditional distributions whose PIT histograms are close to uniform and whose empirical coverage closely tracks nominal levels (Section~\ref{res1:calibration}), indicating that the learned $\hat{p}(\mathbf{y}{\mid}\mathbf{c})$ places probability mass in the right regions of outcome space, including low-density tails that drive extreme centiles. 

Importantly, the diffusion framework is based on sampling: we can draw arbitrary number of conditional samples (e.g., for our 20-IDP SAINT model, we were able to draw $200{,}000$ samples in approximately 2 minutes) and construct empirical CDFs and centiles without committing to specific low-dimensional parametric summaries (e.g., mean and variance or a small set of shape parameters for each IDP, fitted independently). Deviation scores are therefore defined with respect to the full conditional distribution, preserving asymmetry, multimodality and cross-feature structure that may be compressed or distorted by low-dimensional parametric approximations.

\subsection{Joint modeling and multivariate dependence}

A central contribution of our work is demonstrating that diffusion models can jointly model dozens to hundreds of IDPs while preserving realistic dependence structure. Most normative modeling applications in neuroimaging adopt a univariate strategy, fitting a separate model for each imaging phenotype/feature \citep{rutherford2022_normative_protocol,rutherford2022_brain_charts_high_precision}; hence, each IDP is conditionally independent given covariates. While this simplifies computation, it discards potentially informative relationships between regional anatomic measurements that reflect underlying neurobiology.

Our pairwise joint distribution evaluation (Section~\ref{sec:dep-structure}) provides empirical evidence that joint diffusion models capture dependence beyond what can be recovered from marginals alone. Across neurodegeneration-relevant IDPs --- hippocampus, amygdala, ventricular volumes, and AD-related cortical thickness --- the generated pairwise joint distributions systematically depart from a product-of-marginals baseline and better match held-out pairwise joints under MMD and energy distances. These dependencies reflect coordinated atrophy patterns that unfold across structurally and functionally connected regions \citep{Frisoni2010,Dickerson2009}. Capturing higher order structure enables information from the composite deviation profiles to be used in a way that may be more sensitive to early pathology than univariate $z$-scores derived from independent models.

The pair-of-pair analysis extends this finding to the overall pairwise structure. Strong concordance between real and generated density-shape correlation matrices (Mantel $r{\approx}0.985$ for MLP and $r{\approx}0.991$ for SAINT) indicates that the diffusion model learns not only individual pairwise joint distributions but also how those joints are organised across the full IDP set. This is valuable for constructing data-driven biomarkers that integrate information across multiple regions in a biologically coherent manner. By contrast, approaches that aggregate univariate deviation scores without modeling the structure risk combining measurements in ways that do not correspond to a biologically plausible conditional joint distribution.

Recent work in normative modeling has begun to explore multivariate approaches through VAEs \citep{kumar2023multimodal,aguila2023}, typically in moderate-dimensional settings ($D{\le}20$) or following dimensionality reduction. Our results demonstrate that diffusion models with appropriate backbones can scale to higher dimensions ($D = 100$, $200$) while maintaining calibration and dependence structure, opening pathways to normative modeling that encapsulates more of the available imaging information.

\subsection{Architectural considerations: MLP versus SAINT}

The choice of denoiser backbone critically determines performance in high-dimensional regimes. Our experiments reveal a clear distinction between MLP and SAINT architectures: while both performed comparably when the dimensionality is modest (e.g., $D{\le}20$), the MLP backbone degraded markedly at dimension of 100 and 200, exhibiting saturated coverage (empirical coverage probability near 1.0 across nominal intervals), collapsed PIT distributions (mass concentrated near $u{\approx}0.5$), and high KS rejection fractions (Section~\ref{res1:calibration}--\ref{res:distributional_divergence}). By contrast, SAINT maintained acceptable, though slightly degraded, calibration and distributional fidelity even in these settings.

This performance gap stems from fundamental architectural differences. Standard MLPs process features as a single concatenated vector, requiring the network to learn all pairwise and higher-order feature interactions through fully-connected layers. As dimensionality increases, the number of potential interactions grows combinatorially, overwhelming the capacity of fixed-width MLPs. Additionally, MLPs lack mechanisms to distinguish which features are most informative, forcing the network to allocate representational capacity uniformly across all inputs.

SAINT addresses both limitations through its hybrid attention architecture \citep{saint_trl}. Self-attention layers enable the model to selectively weight features within each sample, focusing representational capacity on the most informative IDPs while down-weighting noisy or redundant measurements. In addition, our SAINT-style denoiser includes a lightweight row-attention block that, during training, applies multi-head attention to per-row summary tokens across the mini-batch. This intersample attention encourages the network to relate each subject to other subjects in the same batch, which is loosely analogous to a learned, soft nearest-neighbour mechanism on the local batch rather than an explicit global $k$-NN search \citep{saint_trl}. In our implementation we disable true intersample attention at test time to avoid batch-dependent predictions, so any “borrowing of strength” across individuals acts indirectly through shared parameters that have been shaped by intersample attention during training. Empirically, including this row-attention block improves calibration and distributional metrics in high-dimensional settings, consistent with the view that it stabilises the denoiser by encouraging it to exploit structure shared across subjects when fitting complex conditional distributions.

The computational cost of feature-wise attention scales quadratically with sequence length, but in our tabular setting the "sequence" corresponds to the number of features $D$, which remains tractable on modern hardware. Training times for SAINT on our largest model ($D{=}200$) remained within practical limits (i.e., less than 2 hours, see Table~\ref{tab:runtimes}). Taken together, these results suggest that attention-based denoisers, rather than plain MLPs, should be the default choice when constructing high-dimensional diffusion-based normative models for neuroimaging IDPs.

\subsection{Comparison to traditional normative methods}

GAMLSS and related distributional regression frameworks remain widely used for brain-chart construction due to their interpretability, computational efficiency, and mature software ecosystems \citep{rigby2005gamlss,Bethlehem2022}. These methods model distributional parameters (location, scale, shape) as smooth functions of covariates under flexible but parametric distributions and provide direct centile estimates and uncertainty intervals. In our experiments, the GAMLSS baseline achieved competitive ACE and KS performance for the subset of UKB IDPs where the GAMLSS models converged, and behaved similarly to the diffusion backbones on the SYNTH dataset. However, the GAMLSS fitting procedure failed to converge for nearly half (9/20) of the UKB IDPs, highlighting practical robustness issues when scaling univariate parametric models to comprehensive phenotype datasets.

Because standard GAMLSS is formulated as a univariate distributional regression framework, normative modeling studies commonly fit one model per IDP/phenotype \citep{bozek_2023,Bethlehem2022}, thereby precluding an explicit dependence structure. Attempts to capture multivariate relationships post hoc --- e.g., by correlating univariate $z$-scores that have been marginally standardised --- may neglect the conditional dependence given covariates and can lead to misleading inferences about joint abnormality across IDPs. Moreover, parametric families impose strong distributional assumptions that may be violated in practice. While GAMLSS offers a catalogue of distributions, model selection becomes unwieldy when fitting many univariate models, and misspecification in any single model can distort deviation scores.

Diffusion models occupy a distinct position in this landscape. They combine the flexibility to model complex conditional densities with training dynamics that are typically stable and do not require delicate hyperparameter tuning. They naturally extend to joint multivariate distributions without architectural changes: the same denoising objective applies whether the dimension is 1 or 200. Computational costs grow approximately linearly with the number of model parameters and diffusion timesteps, and for attention-based backbones additionally with the square of the IDP dimension, but modern GPUs make both training and conditional sampling tractable even for high-dimensional settings (e.g., training our 200-IDP SAINT model on UKB data required less than 2 hours runtime). Beyond centile estimation, the generative nature of diffusion models enables conditional synthesis of realistic multivariate IDP profiles, supporting downstream applications such as training-set augmentation for rare deviation patterns, generation of privacy-preserving surrogate cohorts, and systematic stress-testing or validation of statistical pipelines under controlled, data-driven perturbations.

The primary trade-off is interpretability. Parametric centile charts yield closed-form distributional summaries, whereas diffusion-derived centiles emerge from empirical CDFs of conditional samples. Where high-dimensional joint structure and generative flexibility are central, diffusion-based normative models provide a compelling alternative.

\subsection{Clinical implications and personalised biomarkers}

Normative modeling ultimately aims to support precision medicine by quantifying individual-level deviations from healthy reference ranges \citep{rutherford}. Our framework extends this paradigm by enabling deviation profiles that respect multivariate brain structural organisation. Rather than flagging a quantity such as hippocampal atrophy in isolation, a clinician could assess whether an individual's joint pattern of hippocampal volume, ventricular enlargement, and cortical thinning across AD-signature regions \citep{Dickerson2009} collectively deviates from the normative distribution. Such profiles may improve the sensitivity of early disease detection by integrating changes that are individually sub-threshold across multiple biomarkers.

The generative nature of diffusion models also facilitates counterfactual reasoning. Given an individual's observed measurements $\mathbf{y}_{\text{obs}}$ and covariates $\mathbf{c}_{\text{obs}}$, one can generate samples from the normative distribution $\hat{p}(\mathbf{y} \mid \mathbf{c}_{\text{obs}})$ to visualise the range of typical variation, aiding interpretation of $\mathbf{y}_{\text{obs}}$ in context. Alternatively, conditioning on a subset of IDPs (e.g., this could be achieved by taking a large-N sample and restricting this to a subset of IDPs contained in a particular neighbourhood) and sampling the remaining features allows exploration of how abnormality in one region constrains expectations for others under the learned dependence structure.

An important consideration for clinical translation is uncertainty quantification. Diffusion models provide predictive distributions through conditional sampling, enabling credible intervals and coverage probabilities for individual predictions. Our calibration analyses (Section~\ref{res1:calibration}) demonstrate that these intervals achieve target coverage rates, ensuring that stated confidence levels are empirically justified. In high-stakes clinical decisions where quantifying uncertainty is important, diffusion-based normative models can provide this in a relatively straightforward manner. Nonetheless, translating diffusion-based normative models into clinical tools will require further validation in disease cohorts, evaluation of prognostic value, and development of interpretable visualisations of multivariate deviation profiles.

\subsection{Limitations}

Our evaluation of pairwise dependence structure focuses on second-order statistics (MMD, energy distance) and pair-of-pair density-shape correlation. These measures are informative but do not capture all aspects of higher-order dependence. Alternative approaches based on mutual information, copula-based dependence, or topological data analysis might reveal discrepancies between real and generated distributions in higher dimensions.

A second, more general, limitation is that our diffusion-based normative models inherit the properties and biases of the reference cohort on which they are trained, in the same way as existing parametric and hierarchical normative approaches. The modelling framework itself is agnostic to cohort composition and can, in principle, be applied to any suitably large dataset; however, the resulting centiles and deviation scores are only as representative as the underlying sample. In the present work we use the UK Biobank imaging cohort, which is known to exhibit healthy volunteer and socioeconomic selection biases: participants are typically healthier and more socioeconomically advantaged than the general population \citep{Fry2017}. Normative ranges derived from UKB may therefore misrepresent individuals from lower socioeconomic strata, underrepresented ethnic groups, or populations with higher comorbidity burdens. This caveat is particularly relevant for neurodegenerative applications, where the prevalence, age-of-onset, and expression of pathology can vary substantially across demographic strata. Mitigating this limitation requires either training models on cohorts that are more representative of the target clinical population, or using domain-adaptation, transfer-learning or alternative conditioning strategies to adapt UKB-trained models to new sites and populations \citep{BAYER2022,Kia2022Closing}. In this sense, any mismatch between diffusion-based normative estimates and the true target-population distribution should be attributed primarily to cohort selection and sampling bias rather than to intrinsic properties of the modelling framework.

\subsection{Future directions}

While our results show that diffusion-based normative models can provide calibrated, high-dimensional conditional densities in large population cohorts, translating these methods into useful clinical tools will require further development. Here, we outline future directions that we see as relevant to this aim.

First, our experiments have focused on models trained on population-based cohorts and have been evaluated primarily on distributional calibration and dependence structure. A natural next step is to test whether deviation maps derived from diffusion-based normative models add predictive or diagnostic value in pathological cohorts. This may include assessing whether multivariate deviation profiles improve discrimination between patient subgroups, predict conversion or progression, or assist in stratifying heterogeneous clinical populations relative to existing normative model baselines.

Future work should also address demographic and sampling biases, particularly those inherent in UKB \citep{Fry2017}. Strategies such as inverse-probability reweighting, domain adaptation, or federated learning may help adapt models to more representative clinical samples. In addition, extending the framework to longitudinal data, multi-modal imaging (e.g., combining structural, diffusion, and functional measures), and non-imaging covariates (genetic, cognitive, lifestyle) could yield richer multivariate normative models.

Finally, clinical adoption will likely depend on interpretability and workflow integration. Diffusion-based normative models can return samples from a high-dimensional conditional density, which require further processing to convert them into a useful output for clinicians or other end-users. Future work should therefore develop interpretability tools tailored to normative modeling, including: (i) visualisations of joint deviation patterns across anatomically organised IDPs, (ii) counterfactual sampling methods that identify which measurements drive an individual's abnormality score, and (iii) uncertainty summaries that clearly convey model confidence. Embedding these tools in prototype decision-support interfaces and evaluating them with clinicians may help to identify suitable use-cases, understand failure modes, and clarify how diffusion-based normative models can complement existing radiological and clinical assessments.

\subsection{Conclusion}

We have presented a framework for normative modeling in neuroimaging based on denoising diffusion probabilistic models. Our results demonstrate that diffusion models provide flexible, well-calibrated conditional density estimators capable of modeling low- or high-dimensional imaging-derived phenotypes while preserving multivariate dependence structure. The SAINT transformer backbone, with its hybrid self-attention and intersample attention mechanisms, scales effectively to settings with hundreds of features where standard MLPs fail. Evaluation across synthetic and real data confirms that diffusion-derived centiles achieve target coverage rates, maintain distributional fidelity, and capture realistic pairwise relationships among brain measurements.

By moving beyond univariate normative models, diffusion approaches open pathways to composite deviation biomarkers that integrate information across anatomically-connected regions in a biologically coherent manner. The generative nature of these models further enables synthetic data generation, counterfactual reasoning, and flexible incorporation of additional covariates or modalities, extending the scope of normative modeling beyond centile estimation alone. The foundation established here --- demonstrating that diffusion models can deliver calibrated, high-dimensional joint normative estimates --- provides a starting point for these future efforts aimed at early detection, individualised risk stratification, and a more detailed understanding of neurodegenerative disease.

\section*{Acknowledgements}

\ifblindreview
\else
\noindent This research has been conducted using the UK Biobank Resource under application number 417033. LW is funded by an Australian government Research Training Program (RTP) scholarship and an Adelaide University research supplementary scholarship; and is supported by Adelaide University. LJP and MJ are funded and supported by Adelaide University.
\fi

\section*{Data availability}

\noindent UK Biobank data are available through application to UK Biobank under the terms of access for approved researchers.
\ifblindreview
\else
The specific data used in this study were accessed under application number 417033.
\fi
Restrictions apply to the availability of these data, which are not publicly available.


\section*{Code availability}
\noindent \codeavailability Model weights and any derived data products are shared only to the extent permitted by UK Biobank terms; the codebase is provided independently of UK Biobank data access.



\bibliographystyle{elsarticle-harv}
\bibliography{citation}


\clearpage
\appendix

\section{Supplementary results}
\begin{figure}[htpb]\centering
\includegraphics[width=0.95\textwidth]{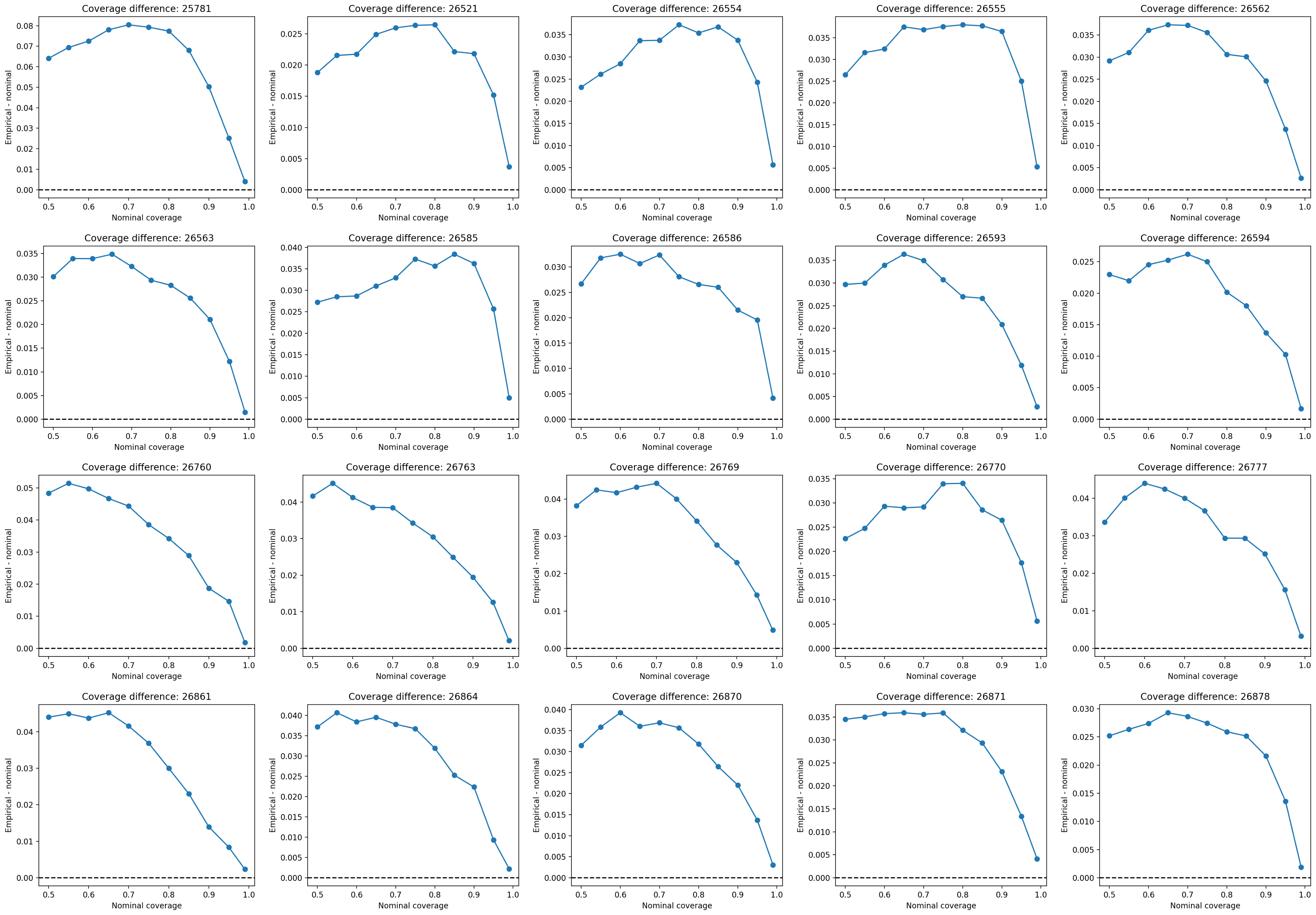}
\caption{Per-IDP coverage difference panels (UKB, MLP diffusion model, base).}
\label{supp:covdiff-peridp-mlp}
\end{figure}
\begin{figure}[htpb]\centering
\includegraphics[width=0.95\textwidth]{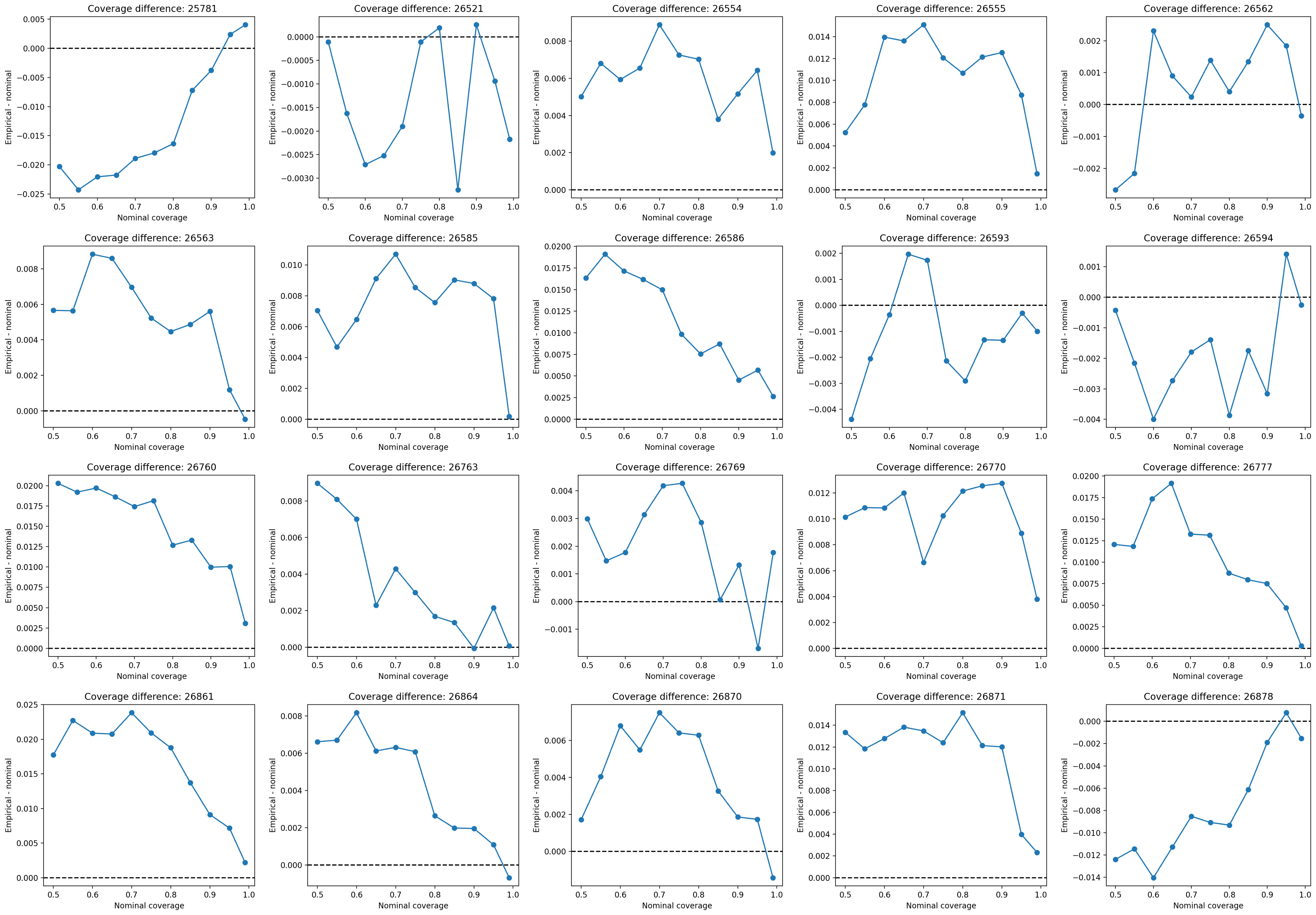}
\caption{Per-IDP coverage difference panels (UKB, SAINT diffusion model, base).}
\label{supp:covdiff-peridp-saint}
\end{figure}
\begin{figure}[htpb]\centering
\includegraphics[width=0.95\textwidth]{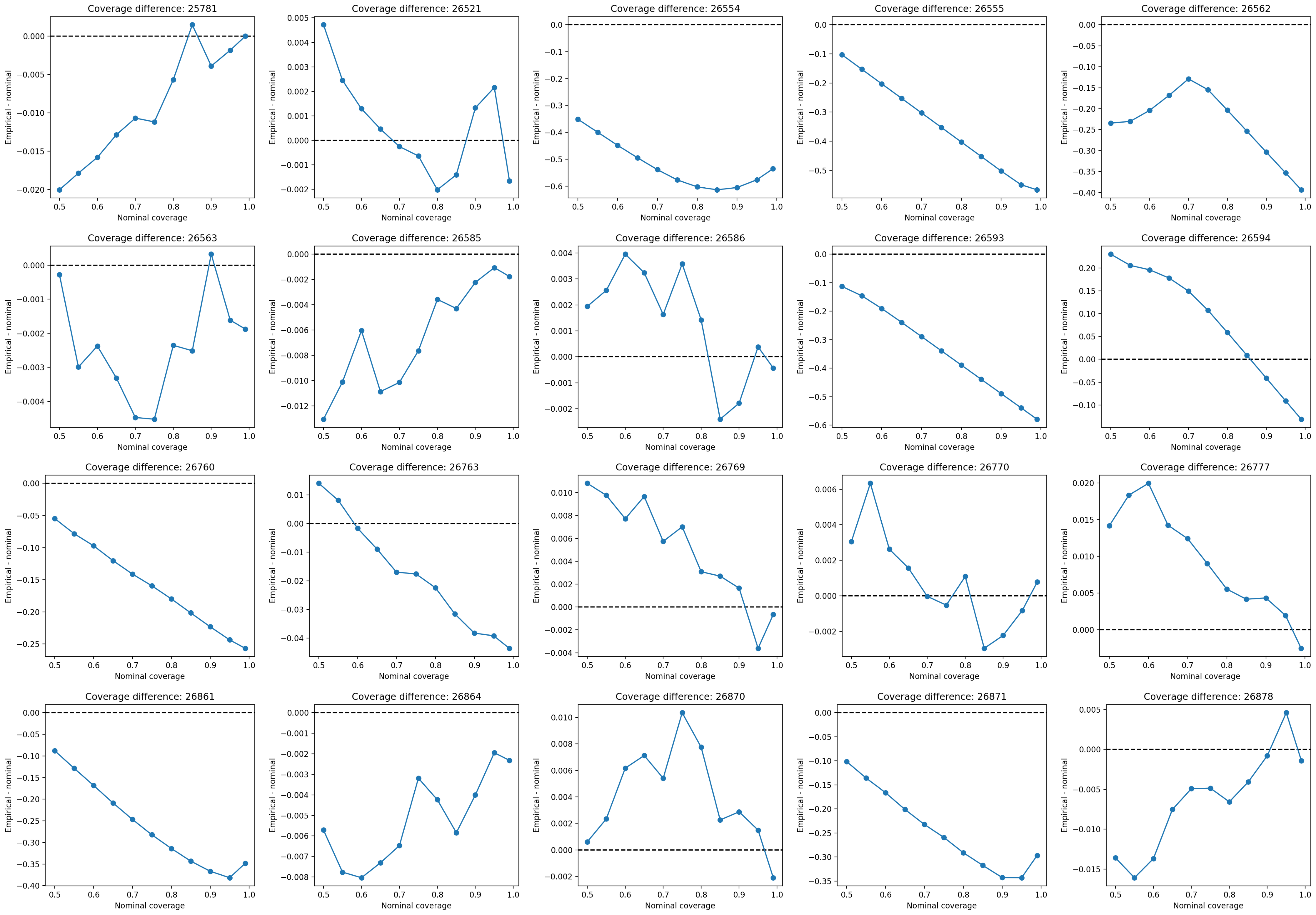}
\caption{Per-IDP coverage difference panels (UKB, GAMLSS baselines, all 20 IDPs).}
\label{supp:covdiff-peridp-gamlss}
\end{figure}
\begin{figure}[htpb]\centering
\includegraphics[width=0.78\textwidth]{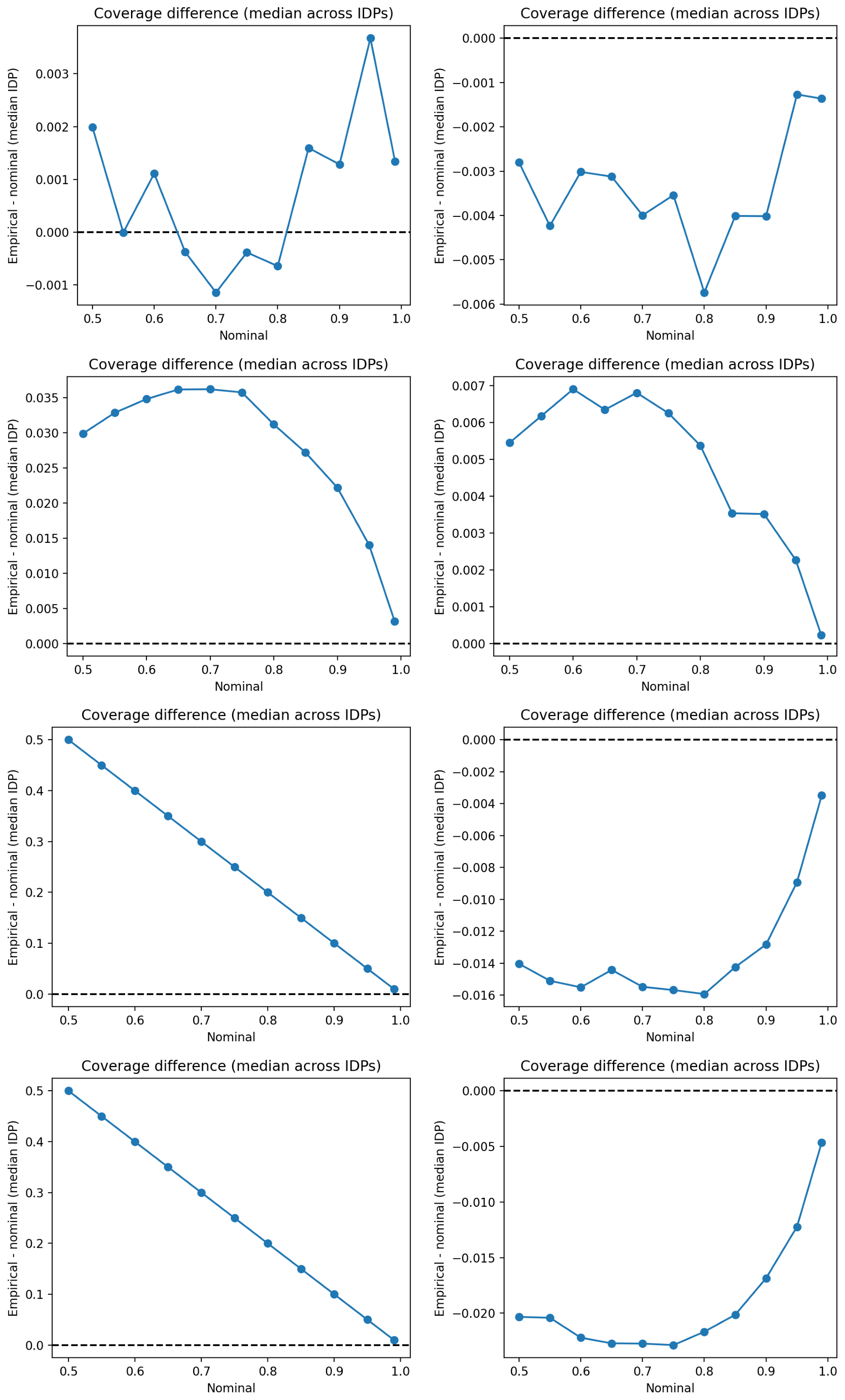}
\caption{Coverage difference median curves for dimensional scaling (UKB diffusion models, MLP/SAINT).}
\label{supp:covdiff-med-dim}
\end{figure}
\begin{figure}[htpb]\centering
\includegraphics[width=0.78\textwidth]{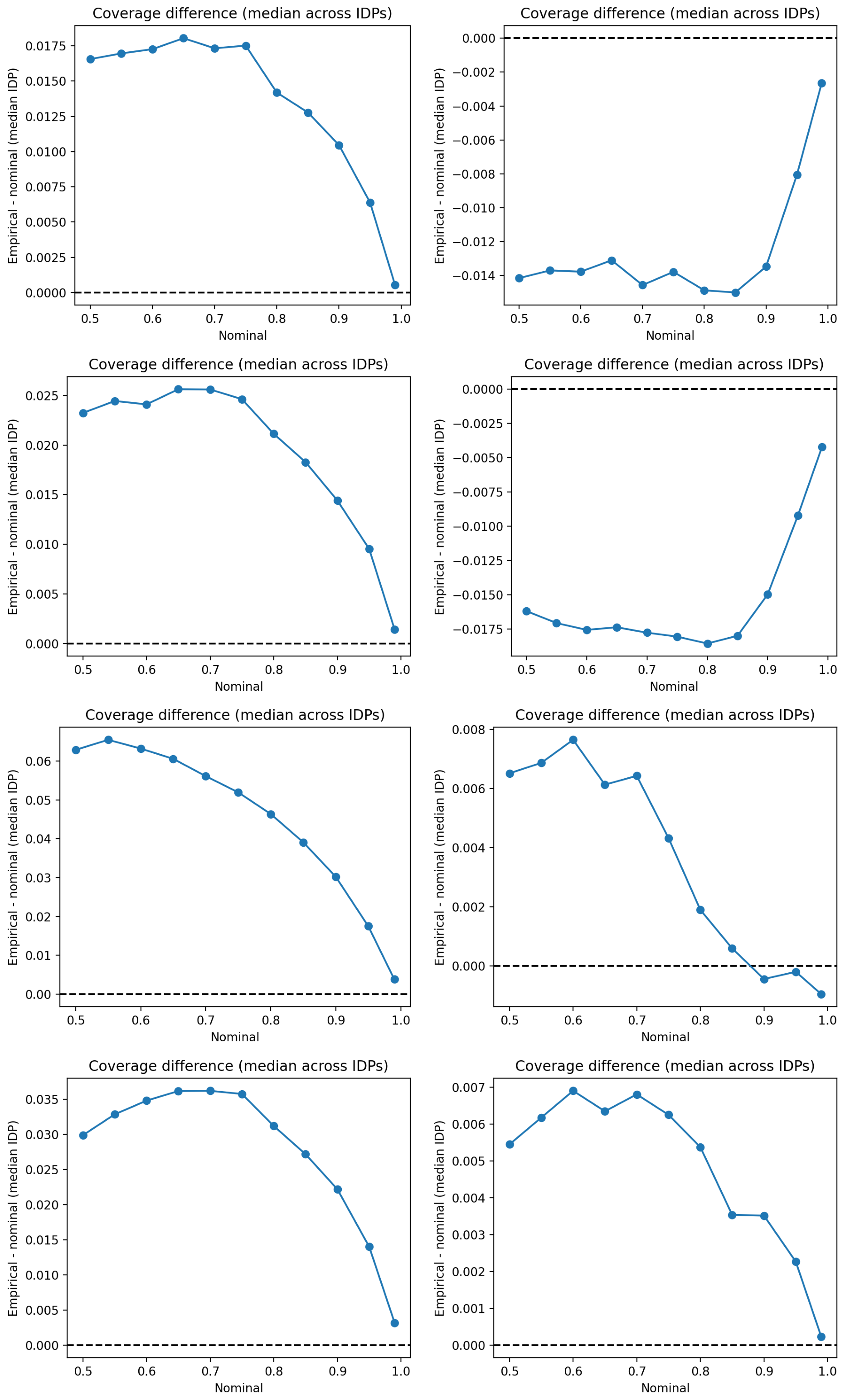}
\caption{Coverage difference median curves for training fractions (UKB diffusion models, D=20).}
\label{supp:covdiff-med-frac}
\end{figure}
\begin{figure}[htpb]\centering
\includegraphics[width=0.95\textwidth]{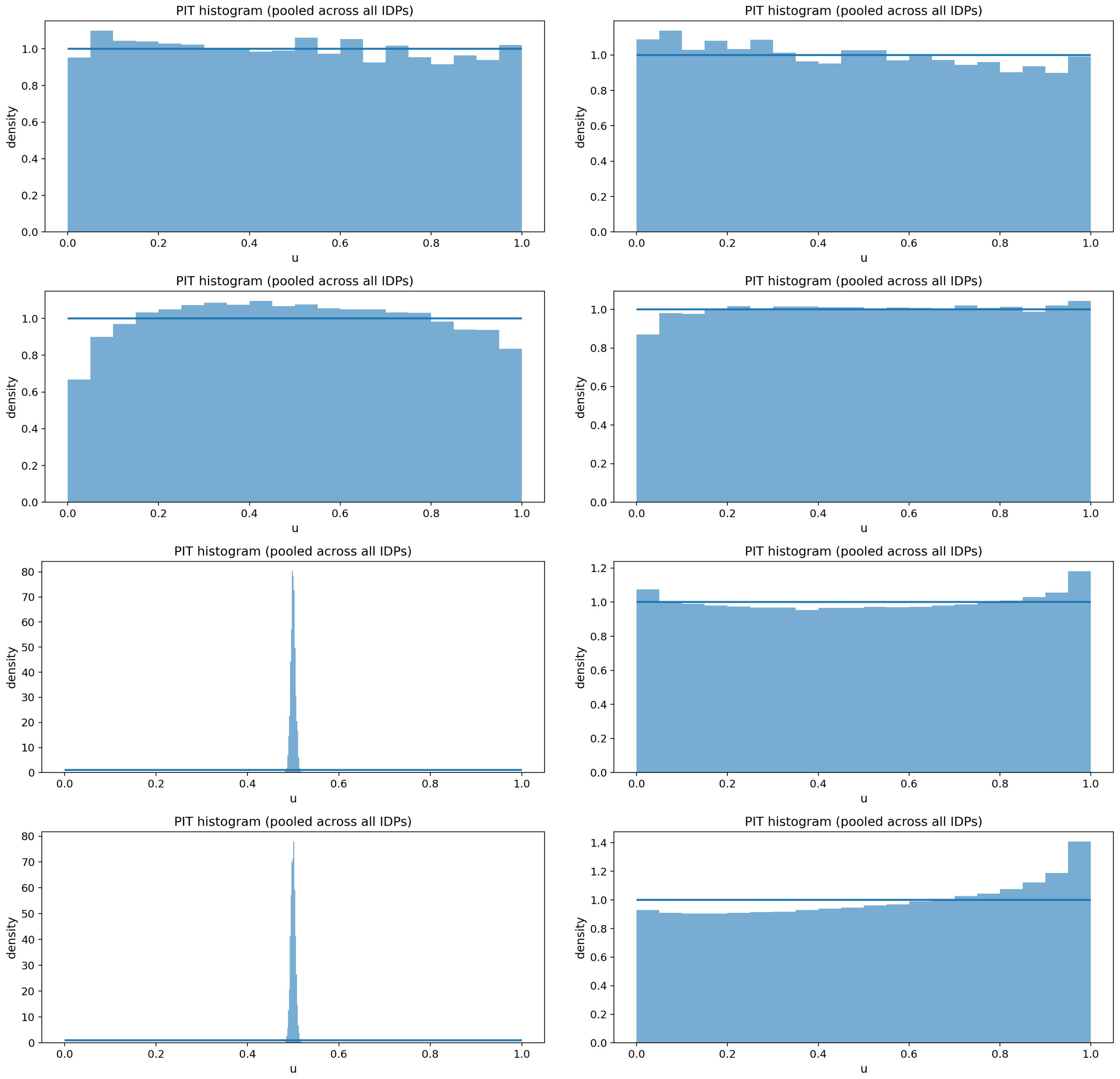}
\caption{Pooled PIT histograms for dimensional scaling (UKB diffusion models, MLP/SAINT).}
\label{supp:pit-dim}
\end{figure}
\begin{figure}[htpb]\centering
\includegraphics[width=0.95\textwidth]{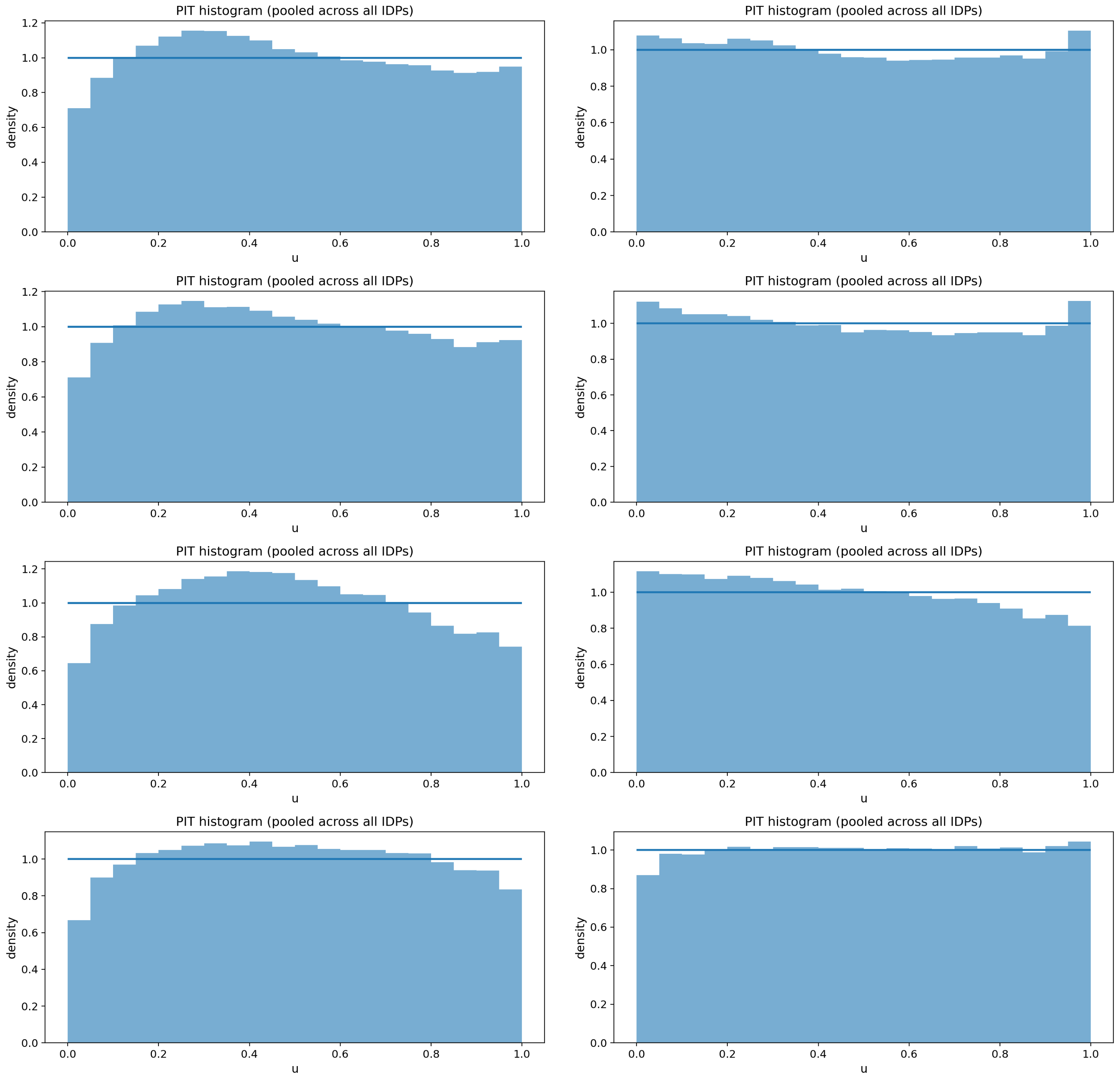}
\caption{Pooled PIT histograms for training fractions (UKB diffusion models, D=20).}
\label{supp:pit-frac}
\end{figure}


\end{document}